\documentclass[10pt,twocolumn,letterpaper]{article}
\usepackage{cvpr}
\usepackage{graphicx}
\usepackage{amsmath}
\usepackage{amssymb}
\usepackage{booktabs}
\usepackage{multirow}
\usepackage{wrapfig}
\usepackage{tabularx}
\usepackage{amssymb}
\usepackage{pifont}
\usepackage[table]{xcolor}
\usepackage{makecell}
\usepackage{float}
\usepackage{adjustbox}
\usepackage[accsupp]{axessibility}  %
\usepackage{color}
\definecolor{bronze}{rgb}{1,1,0.6}
\definecolor{silve}{rgb}{0.969,0.796,0.600}
\definecolor{gold}{rgb}{0.941,0.592,0.600}

\newcommand{\cmark}{\textcolor{green}{\checkmark}}
\newcommand{\xmark}{\textcolor{red}{\ding{55}}}

\newcommand{\methodname}[0]{\textit{AvatarDynamizer}}
\newcommand{\datasetname}[0]{\textit{DynaHuman}}

\newcommand{\best}[1]{\cellcolor{green!25}{#1}}
\newcommand{\second}[1]{\cellcolor{orange!30}{#1}}

\newcommand{\besttext}[1]{\colorbox{green!25}{#1}}
\newcommand{\secondtext}[1]{\colorbox{orange!30}{#1}}

\definecolor{cvprblue}{rgb}{0.21,0.49,0.74}
\usepackage[pagebackref,breaklinks,colorlinks,citecolor=cvprblue]{hyperref}

\begin{document}

\title{AvatarDynamizer: From Static to Dynamic
Human Avatars via \\ Generative Dynamic Textures}

\author{ 
    Guoxing Sun\textsuperscript{1},
    Heming Zhu\textsuperscript{1},
    Linjie Lyu\textsuperscript{1},
    Pascal Fua\textsuperscript{3},
    Christian Theobalt\textsuperscript{1,2},
    Marc Habermann\textsuperscript{1,2}  
    \\
\normalsize{\textsuperscript{1}Max Planck Institute for Informatics, Saarland Informatics Campus}
    \quad
    \normalsize{\textsuperscript{2}VIA Research Center}
    \quad
    \normalsize{\textsuperscript{3}EPFL}\\
    \normalsize{\{gsun, hezhu, llyu, theobalt, mhaberma\}@mpi-inf.mpg.de \quad pascal.fua@epfl.ch}\\
    \normalsize{\url{https://vcai.mpi-inf.mpg.de/projects/AvatarDynamizer/}}
}

\twocolumn[{%
\renewcommand\twocolumn[1][]{#1}%
\maketitle
\vspace{-24pt}
\begin{center}
  \centering
  \includegraphics[width=0.97 \linewidth]{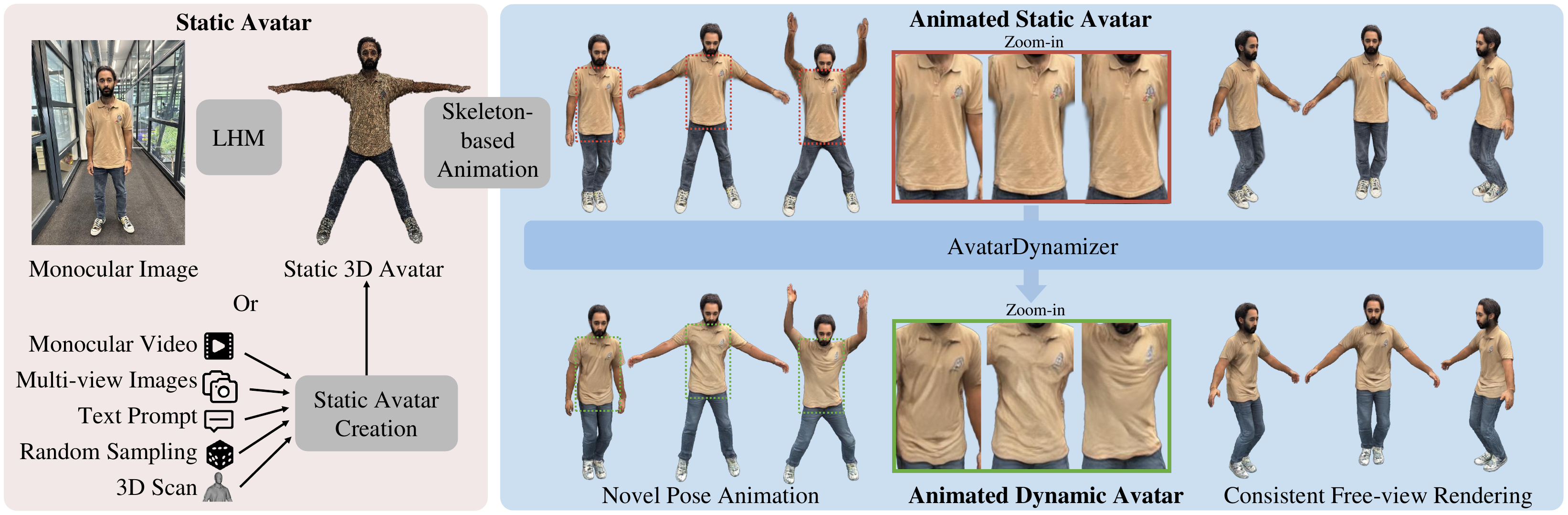}
  \vspace{-10pt}
  \captionof{figure}{
        Given a 3D static avatar from any multimodal input, e.g., a monocular image with LHM~\cite{qiu2025lhm}, we introduce a novel method, \methodname{}, which converts static avatars into controllable 4D human avatars with pose-dependent surface dynamics (see the shirt).
        }
    \label{fig:futureTeas}
\end{center}
\vspace{-5pt}
}]

\begin{abstract}
    For full-body avatars, modeling surface dynamics is crucial for overcoming the uncanny valley and achieving perceptual realism.
Person-agnostic methods recover \textit{static} 3D avatars from monocular images, videos, or text prompts, but their skeleton-driven animations lack realistic surface dynamics such as clothing wrinkles.
In contrast, person-specific methods achieve high-quality rendering and realistic dynamics, but require expensive multi-view captures for each individual.
Recent generalizable dynamic avatar methods struggle to embed surface dynamics, leading to either limited multi-view consistency or dynamic expressiveness.
To this end, we propose \methodname{}, a generative method that transforms an off-the-shelf static 3D avatar into a controllable, realistic, and multi-view-consistent 4D avatar.
We introduce a novel texture-space surface-dynamics embedding and formulate avatar dynamics modeling as conditional texture generation.
Our encoder--decoder representation embeds pose-dependent dynamics into dynamic texture maps, enabling compatibility with pre-trained video diffusion models while decoding them into 3D Gaussians for multi-view consistent rendering.
Since existing datasets are limited in scale, sequence length, or motion diversity, we collect a large-scale multi-view dataset with long sequences covering diverse skeletal motions and surface dynamics.
Experiments show that our method effectively animates static avatars with faithful surface dynamics and outperforms competing generalizable methods in visual fidelity, especially under limited dynamic training data.

\end{abstract}

\section{Introduction} \label{sec:introduction}
Creating faithful and controllable digital twins of real humans has long been of interest in Vision and Graphics, with broad applications in telecommunications, remote coaching, and virtual agents.
Recently, substantial effort has gone into collecting large-scale datasets~\cite{li2025mvhumannet++, cheng2023dna} for generalizable human reconstruction from sparse views~\cite{qiu2025lhm,zubekhin2025giga}, as well as for the generation of 3D avatars from multi-modal inputs such as text~\cite{liao2024tada,dong2024tela} or from unconditional noise~\cite{zhang2024e3gen,chen2023primdiffusion}.
All these methods generate a \textit{static} avatar, which can be animated into arbitrary poses using an underlying rigged skeleton.
However, the resulting animations primarily rely on standard skinning-like deformations~\cite{lewis2000pose} that are approximately piecewise rigid.
Realistic surface dynamics, including cloth wrinkles and induced shading effects, are still missing.
This leads to an uncanny valley effect, making the avatars appear unnatural to human observers.
In this work, we ask: Can we introduce faithful dynamics to off-the-shelf static 3D avatars to bridge this gap?
\par
Traditionally, modeling avatar surface dynamics in Computer Graphics requires manual effort throughout the animation pipeline, including rigging, skinning, animation, cloth simulation, and rendering.
In the machine learning domain, some works learn a mapping from body motion to 4D geometry and appearance represented as radiance/SDF fields~\cite{zhu2024trihuman}, Gaussian splats~\cite{Pang_2024_CVPR,junkawitsch2025eva}, light fields~\cite{kwon2023deliffas}, or conventional 2D textures~\cite{li2024animatable}.
To date, these methods have achieved unprecedented rendering quality and fine-grained control.
However, each character must be captured with a multi-camera system, after which a person-specific model must be trained for several days.
Moreover, they remain person-specific and do not generalize across identities.
\par
Recently, research has shifted toward generalizable approaches~\cite{xu2023magicanimate, chang2023magicpose, zhu2024champ, wang2026wananimate2}, which leverage 2D foundation models~\cite{rombach2022high} to generate dynamic images from an identity image and 2D pose controls.
However, such 2D generation cannot produce multi-view-consistent, free-view renderings.
Recent works further explore cross-identity generalization together with 3D surface dynamics modeling~\cite{guo2025vid2avatar,lu2025gas,shao2024human4dit}.
We observe that their key difference lies in where surface dynamics are embedded.
Image-space methods~\cite{lu2025gas,shao2024human4dit} embed dynamic surface details into video priors and 3D pose controls, enabling them to leverage pre-trained foundation models but sacrificing 3D consistency.
In contrast, texel- or Gaussian-regression methods~\cite{guo2025vid2avatar} embed surface dynamics into a deterministic feed-forward regression framework, directly mapping identity and motion conditions to avatar representations.
While this formulation naturally preserves 3D consistency, it prevents them from directly exploiting pre-trained 2D video generative priors.
Moreover, deterministic regression struggles with modeling fine-grained surface dynamics under limited data, as motion-dependent wrinkles are high-frequency and partially stochastic: the same identity and pose may correspond to different plausible cloth states depending on unobserved physical conditions.
Thus, regression-based methods can easily produce over-smoothed or nearly static appearance dynamics, even when motion conditions are provided.
\par
These observations suggest that the key bottleneck lies in how surface dynamics are represented and embedded. 
A desired representation should satisfy two properties: it should be temporally and spatially consistent, while remaining compatible with powerful generative models such as image and video diffusion models.
To this end, inspired by latent diffusion~\cite{rombach2022high}, we introduce a novel texture-space surface dynamics embedding within an encoder--decoder human representation.
We explicitly embed pose-dependent surface dynamics into colored dynamic texture maps, rather than modeling them via deterministic regression or direct image-space generation.
These dynamic textures serve as a generative intermediate representation: they are spatially and temporally coherent in texel space, compatible with pre-trained video diffusion models, and can be decoded into 3D Gaussians for multi-view consistent rendering.
\par
More precisely, given a static 3D human avatar from diverse multi-modal sources, such as a monocular image, text prompt, 3D scan, or multi-view images, we aim to build a generalizable avatar dynamics model, dubbed \methodname{}, which transforms the static avatar into a controllable 4D avatar with faithful surface and appearance dynamics (see also Fig.~\ref{fig:futureTeas}). 
At its core, our method learns to generate motion-conditioned dynamic texture maps that explicitly embed pose-dependent surface dynamics.
\par
Our method consists of three steps: 1) multi-view data encoding and \textit{Generalized Gaussian Decoder} learning, 2) dynamics generator learning, and 3) inference with a novel static avatar.
We first encode multi-view videos into 2D texture videos through multi-stage optimization.
Here, we non-rigidly register SMPL-X~\cite{SMPL-X:2019} to each frame and then optimize colored 2D dynamic texture maps.
Next, we train our cross-identity \textit{Generalized Gaussian Decoder} to take these texture maps and pose-related normal maps as input and to decode them into 3D Gaussian splats, which are rendered and supervised using multi-view images.
As this step does not require temporal data, any type of multi-view human data can be used to ensure identity generalization.

To learn dynamics for the second step, multi-view video data featuring diverse motions is required, but publicly available data is scarce.
Thus, we collected our own dynamic human dataset, \datasetname{}, comprising 58 actors captured with 100 4K cameras, where each sequence has a total length of $27{,}000$ frames.
While significantly more diverse than any other publicly available dataset, it is still insufficient for full generalization. 
Therefore, we introduce our \textit{Dynamic Texture Generator}, which formulates dynamics learning as a conditional image-to-image translation task in texel space, i.e., given a static color texture and posed normal maps encoding motion, generate the dynamic texture map with pose-dependent wrinkles.
Importantly, it can leverage pre-trained foundational video models~\cite{blattmann2023stable} to compensate for data scarcity.
Finally, at inference time, our method recovers a corresponding static identity texture from a static 3D avatar.
Our contributions are:
\begin{itemize}
\item A novel texture-space surface-dynamics embedding that represents 4D humans as compact 2D dynamic texture maps, together with a \textit{Generalized Gaussian Decoder} for multi-view consistent rendering.
\item 
A \textit{Dynamic Texture Generator} that leverages pre-trained video diffusion priors to generate highly detailed dynamic textures conditioned on identity and motion.
\item \datasetname{}, a large-scale human dataset with diverse identities, camera views, and substantial motion diversity.
\end{itemize}

\section{Related Work} \label{sec:related}
Here, we review prior works on human representations, human priors, and multi-view human datasets for modeling diverse identities, body motions, and clothing dynamics.
Head avatar methods~\cite{cao20133d,thies2016face2face,zhou2023groomgen,qian2024gaussianavatars,prao2024lite2relight,tran2024voodoo,li2025rgbavatar,zhou2025zero1toa,gerogiannis2025arc2avatar,liu2025lucas,Gao2025Learn2Control,he2025lam,wu2026fastavatar}, which focus on facial expressions, skin materials, and hair modeling, are beyond the scope of this work.
\subsection{Human Representations}
\label{subsec:rw_representation}
Developing effective 3D human representations is a fundamental problem for human modeling, reconstruction, and rendering.
Early methods commonly rely on mesh-based templates, often combined with motion capture techniques~\cite{stoll2011fast,kanazawa2018end}, to improve mocap accuracy~\cite{joo2018total,hedlin2022simple} or capture human performances~\cite{deepcap,guo2021human,shetty2024holoported}.
These representations are efficient and compatible with graphics rendering pipelines, but they struggle to capture fine-grained geometry, appearance details, and complex clothing deformation.
With the development of neural rendering, diverse representations have been explored for human avatars, including neural fields~\cite{saito2019pifu,MetaAvatar:NeurIPS:2021,sun2021neural}, point clouds~\cite{wu2020multi,SkiRT:3DV:2022}, NeRF~\cite{peng2021animatable,kwon2021neural}, NeuS~\cite{zeng2023avatarbooth,sun2024metacap}, 3D Gaussians~\cite{hu2024gauhuman,li2024animatable,jiang2025reperformer,chen2025taoavatar,sun2025real,zhu2026uma} and transformer-based representations~\cite{yu2025humanram}.
These methods achieve high-quality rendering and controllability through rigged skeletons.
However, such 3D representations are usually designed for person-specific reconstruction or animation, making it non-trivial to integrate them with pre-trained 2D generative pipelines for generalizable, pose-dependent surface modeling.
Recent works further investigate texture-based representations for human rendering and reconstruction~\cite{shetty2024holoported,sun2025real,alldieck2019tex2shape}.
GIGA~\cite{zubekhin2025giga} unprojects sparse-view images into a texture map and predicts Gaussians in texel space, but it does not adopt a generative formulation.
MoGA~\cite{dong2025moga} learns an implicit texture latent for human inversion; however, its pose-independent design restricts it to static human generation and prevents effective modeling of dynamic surface changes.
In contrast, we reinterpret the texture map as an explicit latent representation of a 3D human from an encoding--decoding perspective.
This explicit latent is pose-aware, changes smoothly under pose control, and serves as a bridge between pre-trained 2D video generative priors and multi-view consistent 3D avatar rendering.
\subsection{Human Priors}
\label{subsec:rw_prior}
Human priors are widely used to regularize optimization, improve reconstruction, and enable generalizable generation.
In motion capture, prior works use hand-crafted constraints or learned statistical priors~\cite{stoll2011fast,bogo2016keep,SMPL-X:2019,davydov2022adversarial} for robust body pose estimation.
For image-based reconstruction, regression-based methods learn priors to predict neural fields~\cite{saito2019pifu,xiu2022icon,alldieck2022phorhum}, multi-view images~\cite{huang2025adahuman,yu2025humanram}, normal maps~\cite{xiu2023econ,li2025pshuman}, or 3D Gaussians~\cite{zhuang2024idolinstant,qiu2025lhm,qiu2025lhmpp}.
While generalizing across identities and clothing, they mainly target static or frame-wise reconstruction and struggle with temporally coherent, pose-dependent surface dynamics.
Diffusion-based monocular approaches such as AvatarPopUp~\cite{kolotouros2024avatarpopup} exploit generative priors for single-view avatar reconstruction, but controllable 4D modeling with temporal wrinkles and multi-view consistency remains challenging.
Another line of work adopts image or video diffusion models to represent humans in 2D image space~\cite{wang2025meat,xue2025infinihuman,xu2023magicanimate,zhu2024champ,chang2025xdynaexpressivedynamichuman,li2025pshuman,huang2025adahuman}.
These methods benefit from strong identity generalization enabled by large-scale generative priors, but they often lack multi-view consistency.
Toward consistent 4D avatar generation, HDA~\cite{cao2025hyper} learns a generative model in the hyperspace of an animatable avatar representation~\cite{Pang_2024_CVPR}, though its quality is limited by the fitting-based preprocessing of network weights.
Most closely related to our work, Vid2AvatarPro~\cite{guo2025vid2avatar} and GAS~\cite{lu2025gas} aim to model generalizable human surface and appearance dynamics.
Vid2AvatarPro~\cite{guo2025vid2avatar} conditions on both identity and pose for dynamic avatar reconstruction, but jointly learning identity and pose control through a regression model is challenging and often results in static appearances even when pose input is provided.
Moreover, directly decoding 3D Gaussians makes it difficult to leverage pre-trained foundation video models.
GAS~\cite{lu2025gas} learns video generation priors with a switcher to produce dynamic wrinkles for a static 3D avatar in image space, but this comes at the cost of multi-view consistency.
In this work, we embed surface dynamics into texture space: a video diffusion prior models motion-dependent texture dynamics, while a reconstruction prior decodes the generated textures into 3D Gaussians for multi-view consistent rendering.
\subsection{Human Datasets}
\label{subsec:rw_dataset}
High-quality multi-view human datasets are essential for learning generalizable and dynamic human avatars.
Existing datasets mainly fall into two categories, each with distinct limitations for modeling both identity diversity and pose-dependent surface dynamics.
On the one hand, dynamic datasets of individual subjects, such as DDC~\cite{habermann2021} and DUT~\cite{sun2025real}, capture detailed body motions and surface deformations.
These datasets are valuable for learning precise dynamics, but they contain very few identities, which limits cross-subject generalization.
On the other hand, large-scale identity datasets such as MVHumanNet++~\cite{li2025mvhumannet++} and DNA-Rendering~\cite{cheng2023dna} provide diverse human appearances, clothing styles, and body shapes.
However, they generally contain static poses or very limited motion ranges, making them insufficient for learning complex dynamic surface priors such as temporal cloth wrinkling.
This dichotomy motivates us to leverage pre-trained 2D video diffusion priors to bridge the gap between large-scale identity diversity and realistic human dynamics, and to collect a large-scale dataset with diverse identities and body motions.
\begin{figure*}[tb]
    \centering
	\includegraphics[width=0.95\linewidth]{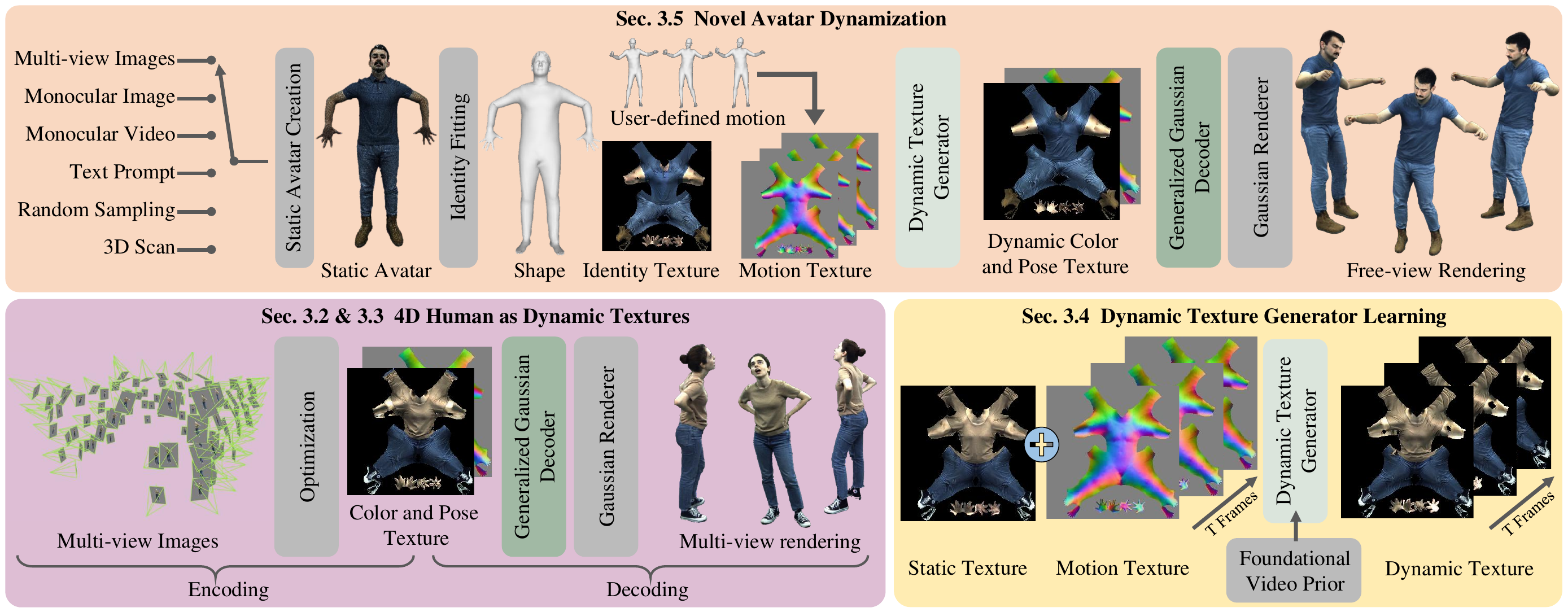}
    \vspace{-5pt}
	\caption
	{
	    \textbf{Overview}.
        Given a static human avatar reconstructed from multimodal inputs, e.g., multi-view images, we first perform identity fitting to obtain its coarse shape and identity texture.
        Our \textit{Dynamic Texture Generator} predicts dynamic texture maps conditioned on identity and novel pose.
        Then, our \textit{Generalized Gaussian Decoder} recovers Gaussian splats for faithful and view-consistent renderings.
	}
	\label{fig:overview}
    \vspace{-15pt}
\end{figure*}

\section{Method} \label{sec:method}
We introduce \methodname{}, which transforms a static 3D human avatar from diverse input modalities, such as monocular or multi-view images, into a controllable 4D avatar with pose-dependent deformations and appearance changes, e.g., shadows and wrinkles (Fig.~\ref{fig:overview}).
Importantly, our renderings are inherently multi-view consistent, as the final dynamic avatar is represented using Gaussian splats.
To improve 4D human encoding, we introduce a novel texture-space human representation that encodes multi-view human videos into compact dynamic textures through multi-stage optimization.
We train our \textit{Generalized Gaussian Decoder} to reconstruct texel-aligned 3D Gaussian splats from dynamic color and pose textures, supervised by multi-view images (Sec.~\ref{subsec:representation}).
With multi-view data embedded, our \textit{Dynamic Texture Generator} formulates dynamics learning as conditional video generation in texture space, using pose-related normal maps and identity textures as inputs (Sec.~\ref{subsec:prior}).
This formulation enables initialization with a pre-trained video foundation model~\cite{blattmann2023stable}, supporting dynamics and identity generalization under limited data while capturing the stochastic nature of surface dynamics.
At inference, we fit the coarse shape and identity texture of the static avatar, generate motion-dependent textures from the identity texture and posed normal map, and decode them into 3D Gaussians for novel-view rendering (Sec.~\ref{subsec:inference}).
Before diving into the method, we review the background (Sec.~\ref{subsec:preliminaries}) and describe the key idea (Sec.~\ref{subsec:idea}).
\subsection{Preliminaries}\label{subsec:preliminaries}
\par \noindent\textbf{Latent Video Diffusion.}
LVD~\cite{blattmann2023stable} learns to generate videos by modeling the reverse diffusion process in a compressed latent space.
A pre-trained 3D variational autoencoder (VAE) $\mathcal{E}(\mathcal{V})=\mathbf{z}_{0}$ first embeds a $T$-frame video sequence $\mathcal{V} \in \mathbb{R}^{T \times H \times W \times 3}$ into a latent $\mathbf{z}_{0}$.
With a predefined noise schedule, we compute a noisy latent $\mathbf{z}_{t}$ at timestep $t$ as $\mathbf{z}_t = \alpha_{t}\mathbf{z}_0 + \sigma_{t}\epsilon$, where $\epsilon \sim \mathcal{N}(0, \mathbf{I})$ denotes the Gaussian noise added to the latent representation.
At each timestep $t$, a denoising network $\epsilon_\theta(\cdot)$ predicts the noise, and the training objective is formulated as:
\begin{equation} \label{eq:lvd}
\mathcal{L}_{LVD} = \mathbb{E}_{\mathbf{z}_0, \epsilon \sim \mathcal{N}(0,\mathbf{I}), t} \left[ \| \epsilon - \epsilon_\theta(\mathbf{z}_t, t, \mathbf{C}) \|^2_2 \right]
\end{equation}
where $\mathbf{C}$ represents the conditioning signals, which could be text prompts, depth maps, or normal maps, and $\theta$ refers to the trainable network weights.
\par \noindent\textbf{Texel Gaussian Avatar.}
We use the parametric model SMPL-X~\cite{SMPL-X:2019} as the base template $\mathbf{\bar{V}}\in \mathbb{R}^{ N_\mathrm{V} \times 3}$ and bind 3D Gaussians to its texture~\cite{Pang_2024_CVPR}.
The coarse shape is defined by the shape parameter $\boldsymbol{\beta}$, cloth displacements $\mathbf{d}_\mathrm{cloth}$ and hair displacements $\mathbf{d}_\mathrm{hair}$.
Given a skeletal motion $\mathcal{M}$, i.e., joint rotations, as input, we animate it through LBS~\cite{lewis2000pose}: 
\begin{equation} \label{eq:template_transformtation}
\mathbf{V} = {T}_{\mathrm{LBS}}(\mathcal{M},{T}_{\mathrm{D}}(\mathbf{\bar{V}}, \boldsymbol{\beta}, \mathbf{d}_\mathrm{hair}, \mathbf{d}_\mathrm{cloth})),
\end{equation}
where we deform the base mesh in the canonical pose using $T_{\mathrm{D}}(\cdot)$.
For each valid texel $i$ of the texture map, we parameterize the corresponding Gaussian as $\mathcal{T}_{\boldsymbol{\mathcal{G}}_{i}} = \{ \boldsymbol{\mu}_{i},\textbf{h}_{i},\textbf{s}_{i},\textbf{r}_{i},\boldsymbol{\alpha}_{i} \}$, including 3D positions in world space, $\boldsymbol{\mu}_{i} = \mathbf{T}_{\mathrm{LBS}}(\mathcal{M},\mathbf{T}_{\mathrm{D}}(\cdot) + \boldsymbol{\delta}_{i}) $, spherical harmonics, scaling, rotations, and opacity values, respectively. 
$\boldsymbol{\delta}_{i}$ denotes a motion-dependent per-Gaussian displacement that is later predicted by our \textit{Generalized Gaussian Decoder} alongside the other Gaussian parameters.
Given a virtual camera view matrix $\mathbf{W}$, Gaussians are rendered into a 2D image with a tile-based 3DGS rasterizer~\cite{kerbl3Dgaussians}: $ \mathcal{R}(\boldsymbol{\mathcal{G}},\mathbf{W}) = \mathcal{\hat{I}}$.
\subsection{Embedding 4D Human as Dynamic Textures} \label{subsec:idea}
\par \noindent\textbf{Motivation.}
For faithful and controllable 4D human avatars, how surface dynamics are represented and embedded is crucial.
Existing representations either lack compatibility with powerful diffusion priors~\cite{guo2025vid2avatar,dong2025moga} or sacrifice multi-view consistency~\cite{xue2025infinihuman,lu2025gas}.
An ideal representation should support generative priors while preserving multi-view consistency.
Inspired by latent diffusion~\cite{rombach2022high}, we embed multi-view images into compact 2D textures, which are decoded by our \textit{Generalized Gaussian Decoder} into texel-aligned 3D Gaussians.
Unlike texel-space methods such as GIGA~\cite{zubekhin2025giga}, where textures serve as sparse-view fusion features for feed-forward reconstruction, we use dynamic textures as an explicit, time-varying latent space for generative surface dynamics.
They are optimized from multi-view videos, generated under motion control, and decoded into 3D Gaussians for consistent rendering.
This shifts texture maps from reconstruction features to a generative dynamics embedding.
To turn a static avatar, represented in our compact texture space, into a dynamic one, our \textit{Dynamic Texture Generator} maps static color textures to motion-dependent dynamic textures conditioned on skeletal motion.
This formulates dynamics generation as conditional video generation in texel space, enabling to leverage video diffusion priors~\cite{rombach2022high,blattmann2023stable} to counter data scarcity.
Unlike image-space methods~\cite{xue2025infinihuman,lu2025gas}, our generated textures are decoded into 3D Gaussians, naturally preserving multi-view consistency.
\par \noindent\textbf{Dynamic Texture Embedding.} 
Embedding multi-view videos into dynamic textures involves multi-stage optimization.
For clarity, we omit the temporal smoothing used during batch optimization.
Given multi-view images $\{\mathcal{I}^{k}\}_{k\in[1,N]}$ captured from $N$ camera views with paired foreground segmentations $\{\mathcal{S}^{k}\}_{k\in[1,N]}$ of each frame, we first reconstruct point cloud geometry $\mathbf{p}$~\cite{neus2} and run 2D marker detection~\cite{cao2019openpose} and triangulation to recover 3D markers $\mathbf{m}$.
With $\mathbf{p}$ and $\mathbf{m}$ as reference, we optimize the pose parameter $\mathcal{M}$ and shape parameter $\boldsymbol{\beta}$ of SMPL-X~\cite{SMPL-X:2019}, yielding coarse geometric tracking.
To capture fine textures, we further optimize time-varying cloth displacements $\tilde{\mathbf{d}}_\mathrm{cloth}$ in the canonical pose on top of the coarsely tracked SMPL-X template, with $\mathbf{p}$ as the reference geometry.
Next, we recover the RGB texture map $\mathcal{T}$ through differentiable rasterization $\mathcal{DR}$~\cite{Laine2020diffrast} from multi-view images:
{\small
\begin{equation}
\min_{\mathcal{T}}
\sum_{k=1}^{N}
\left\|
\hat{\mathcal{I}}_k
-
\mathcal{DR}(\mathbf{V}(\mathcal{M}, \boldsymbol{\beta}, \mathbf{d}_\mathrm{hair}, \tilde{\mathbf{d}}_\mathrm{cloth} ), \mathcal{T}, \mathbf{W_{k}})
\right\|_2^2
\end{equation}
}
Thus, human information is encoded in textures $\mathcal{T}$ for all subjects and frames.
See the supplementary for details.
\subsection{Generalized Gaussian Decoder}
\label{subsec:representation}
With the multi-view data compressed into dynamic textures, the decoder takes them as input to reconstruct 3D Gaussian primitives and render images accordingly.
Specifically, the Generalized Gaussian Decoder $\Phi_\mathrm{GGD}$ takes the texture map $\mathcal{T}$ and the root-normalized normal map $\mathcal{T}_\mathrm{\bar{N}}$ as inputs and predicts the texel-aligned 3D Gaussian parameters
\begin{align}
     \mathcal{T}_{\boldsymbol{\mathcal{G}}} &= \Phi_{\mathrm{GGD}}(\mathcal{T}, \mathcal{T}_\mathrm{\bar{N}}) \; .
\end{align}
We supervise $\Phi_\mathrm{GGD}$ with multi-view image losses, including L1, SSIM~\cite{wang2004image}, and IDMRF~\cite{wang2018image}, and geometric regularization on the displacements of the Gaussians:
{\small
\begin{equation}
    \mathcal{L}_\mathrm{GGD} = \mathcal{L}_\mathrm{L1} + \lambda_\mathrm{SSIM}\mathcal{L}_\mathrm{SSIM} + \lambda_\mathrm{IDM}\mathcal{L}_\mathrm{IDM} + \lambda_\mathrm{Reg}\mathcal{L}_\mathrm{Reg}.
\end{equation}
}
$\Phi_\mathrm{GGD}$ therefore lifts our texture-space surface-dynamics embedding to 3D Gaussians, enabling video diffusion priors to model dynamics in texture space while preserving multi-view consistency after decoding.

\subsection{Dynamic Texture Generator} \label{subsec:prior}
Our texture-space embedding compresses 4D human surface dynamics as dynamic RGB texture videos, providing a compact generative space that can directly leverage pre-trained video diffusion priors.
Next, we introduce the \textit{Dynamic Texture Generator} $\Phi_{\mathrm{DTG}}$ to model motion-dependent surface dynamics directly in texture space.
At this stage, 3D Gaussians are not involved; instead, the generated dynamic textures are later decoded by the \textit{Generalized Gaussian Decoder} for multi-view consistent rendering.
To faithfully model the wrinkles on the dynamic textures, unlike Vid2Avatar-Pro~\cite{guo2025vid2avatar}, which models the prior through regression, we formulate our generator $\Phi_{\mathrm{DTG}}$ as a skeletal motion-conditioned and identity-conditioned video generation task in texel space.
Specifically, we adopt a stable video diffusion model~\cite{blattmann2023stable} as the base model and condition \textbf{character identity} on the RGB texture map $\mathcal{T}_\mathrm{id}$ capturing the character appearance in `A pose', and \textbf{character motion} on a stack of normal maps $\mathcal{T}_\mathrm{\bar{N}}$ rendered from the SMPL-X template mesh with the root translation and rotation factored out.
Under this conditioning scheme,
we formulate the denoising prior model $\Phi_{\mathrm{DTG}}$ as
\begin{equation}
\epsilon_\theta(\mathbf{z}_t, t, \mathbf{C}) = \Phi_{\mathrm{DTG}}(\mathbf{z}_t, t,\mathcal{E}(\mathcal{T}_\mathrm{id}), \mathcal{E}( \{ \mathcal{T}_\mathrm{\bar{N}}^{j} \}_{j\in[1,T]} )),
\end{equation}
which is trained with the denoising loss in Eq.~\ref{eq:lvd} using dynamic textures from multiple subjects.
Once we obtain generated dynamic textures $\{\mathcal{\hat{T}}^{j}\}_{j\in[1,T]}$ that describe the character under the given motions, we feed them into the $\Phi_{\mathrm{GGD}}$ to reconstruct the 4D dynamic Gaussian Splats.
\subsection{Novel Avatar Dynamization}\label{subsec:inference}
Existing methods reconstruct static avatars from text~\cite{liao2024tada}, monocular or multi-view images~\cite{zhuang2024idolinstant,qiu2025lhm,Pang_2024_CVPR}, videos~\cite{guo2023vid2avatar}, or 3D scans~\cite{zheng2022structured}.
We dynamize them in the following steps.
\par \noindent\textbf{Identity Fitting.} 
Given a static avatar, we render multi-view images of it and follow Sec.~\ref{subsec:idea} to optimize rest body pose $\mathcal{M}$, shape parameter $\boldsymbol{\beta}$, cloth displacements $\mathbf{d}_\mathrm{cloth}$, hair displacements $\mathbf{d}_\mathrm{hair}$, and identity texture $\mathcal{T}_\mathrm{id}$.
\begin{figure*}[tb] %
    \centering
	\includegraphics[width=0.95\linewidth]{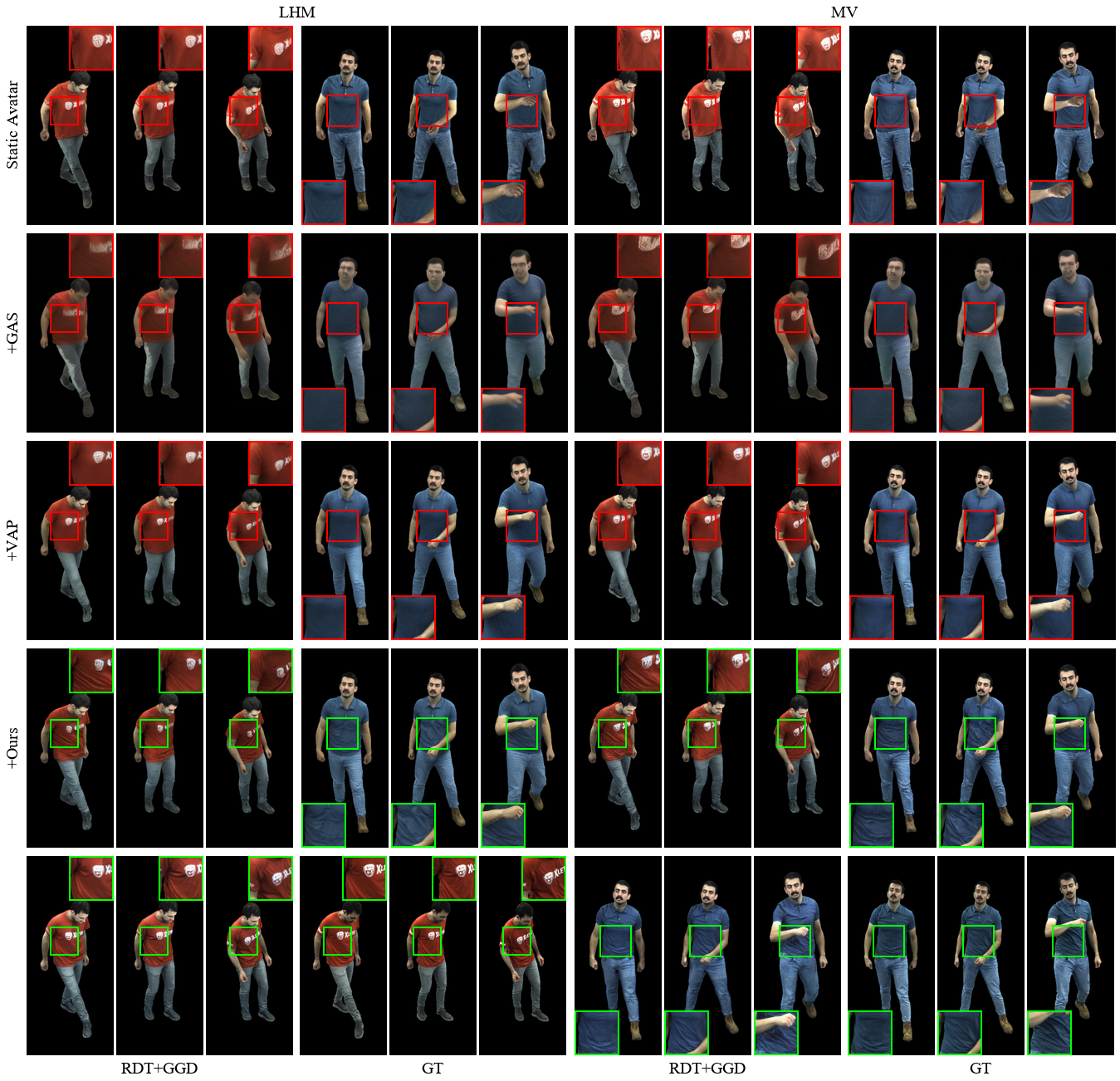}
    \vspace{-12pt}
	\caption
	{
	    \textbf{Qualitative Comparison on DynaHuman}.
        Given a static avatar reconstructed from a monocular image using LHM~\cite{qiu2025lhm} or from multi-view images (MV), we compare our method with state-of-the-art generalizable surface dynamics methods.
        Our AvatarDynamizer produces faithful wrinkles while preserving both 3D geometry and identity consistency.
	}
	\label{fig:comparison}
    \vspace{-15pt}
\end{figure*}

\begin{table*}[t]
    \centering
    \caption{
    \textbf{Quantitative comparison across datasets.}
    Each result is averaged over two subjects.
    Higher PSNR and SSIM are better, while lower LPIPS, FID, and FVD
    are better.
    The \besttext{best} and \secondtext{second-best} results are highlighted
    within each dataset and each method family.
    }
    \label{tab:dataset_comparison_lhm_mv}

    \vspace{-8pt}
    \scriptsize
    \setlength{\tabcolsep}{3.0pt}
    \renewcommand{\arraystretch}{1.13}

    \begin{tabular}{
        @{}
        l
        lccccc
        @{\hspace{8pt}}
        lccccc
        @{}
    }
        \toprule

        \textbf{Dataset}
        & \textbf{Method}
        & PSNR $\uparrow$
        & SSIM $\uparrow$
        & LPIPS $\downarrow$
        & FID $\downarrow$
        & FVD $\downarrow$
        
        & \textbf{Method}
        & PSNR $\uparrow$
        & SSIM $\uparrow$
        & LPIPS $\downarrow$
        & FID $\downarrow$
        & FVD $\downarrow$ \\

        \midrule

        \multirow{4}{*}{\textbf{DeepCap}}
        & LHM
        & 30.61
        & 0.715
        & \second{0.201}
        & \second{50.10}
        & \best{150.12}
        
        & MV
        & 30.86
        & 0.721
        & \second{0.191}
        & \second{39.27}
        & \second{142.18} \\
        
        & +GAS
        & 30.14
        & 0.700
        & 0.248
        & 57.14
        & 223.82
        
        & +GAS
        & 30.48
        & 0.705
        & 0.242
        & 54.28
        & 222.78 \\
        
        & +VAP
        & \best{31.31}
        & \best{0.730}
        & 0.221
        & 101.73
        & 206.88
        
        & +VAP
        & \best{32.64}
        & \best{0.752}
        & 0.197
        & 73.47
        & 178.08 \\
        
        \rowcolor{gray!10}
        & \textbf{+Ours}
        & \second{31.19}
        & \second{0.723}
        & \best{0.200}
        & \best{48.69}
        & \second{160.75}
        
        & \textbf{+Ours}
        & \second{31.95}
        & \second{0.735}
        & \best{0.185}
        & \best{37.05}
        & \best{134.26} \\
        
        \midrule
        
        \multirow{4}{*}{\textbf{DDC}}
        & LHM
        & 29.65
        & 0.696
        & \second{0.206}
        & \second{42.64}
        & \second{108.00}
        
        & MV
        & 29.52
        & 0.698
        & \second{0.189}
        & \second{36.14}
        & \second{103.10} \\
        
        & +GAS
        & 29.40
        & 0.676
        & 0.281
        & 44.32
        & 167.20
        
        & +GAS
        & 29.64
        & 0.680
        & 0.279
        & 44.35
        & 164.61 \\
        
        & +VAP
        & \best{30.81}
        & \best{0.729}
        & 0.224
        & 77.90
        & 199.85
        
        & +VAP
        & \best{31.97}
        & \best{0.750}
        & 0.206
        & 64.51
        & 165.77 \\
        
        \rowcolor{gray!10}
        & \textbf{+Ours}
        & \second{30.70}
        & \second{0.716}
        & \best{0.199}
        & \best{34.97}
        & \best{104.72}
        
        & \textbf{+Ours}
        & \second{31.26}
        & \second{0.728}
        & \best{0.186}
        & \best{30.43}
        & \best{93.95} \\
        
        \midrule
        
        \multirow{4}{*}{\textbf{MetaCap}}
        & LHM
        & 28.11
        & 0.704
        & \best{0.173}
        & \second{80.19}
        & \second{80.19}
        
        & MV
        & 30.06
        & 0.719
        & \best{0.157}
        & \second{23.24}
        & \second{69.69} \\
        
        & +GAS
        & 29.07
        & 0.685
        & 0.275
        & 100.37
        & 100.36
        
        & +GAS
        & 29.33
        & 0.689
        & 0.273
        & 44.50
        & 99.48 \\
        
        & +VAP
        & \best{30.88}
        & \best{0.727}
        & 0.199
        & 115.71
        & 115.71
        
        & +VAP
        & \best{32.28}
        & \best{0.751}
        & 0.177
        & 43.68
        & 106.36 \\
        
        \rowcolor{gray!10}
        & \textbf{+Ours}
        & \second{30.36}
        & \second{0.713}
        & \second{0.177}
        & \best{25.68}
        & \best{60.14}
        
        & \textbf{+Ours}
        & \second{31.46}
        & \second{0.728}
        & \second{0.158}
        & \best{18.31}
        & \best{48.59} \\
        
        \midrule
        
        \multirow{4}{*}{\textbf{DynaHuman}}
        & LHM
        & 25.32
        & 0.719
        & \best{0.193}
        & \second{31.13}
        & \second{75.39}
        
        & MV
        & 25.20
        & 0.727
        & \best{0.181}
        & \second{25.83}
        & \second{52.79} \\
        
        & +GAS
        & 25.74
        & 0.723
        & 0.254
        & 50.03
        & 124.51
        
        & +GAS
        & 25.97
        & 0.725
        & 0.248
        & 49.53
        & 107.15 \\
        
        & +VAP
        & \best{26.81}
        & \best{0.751}
        & 0.211
        & 39.34
        & 79.01
        
        & +VAP
        & \best{27.64}
        & \best{0.771}
        & 0.190
        & 37.51
        & 54.97 \\
        
        \rowcolor{gray!10}
        & \textbf{+Ours}
        & \second{26.43}
        & \second{0.729}
        & \second{0.202}
        & \best{22.58}
        & \best{56.72}
        
        & \textbf{+Ours}
        & \second{26.56}
        & \second{0.736}
        & \second{0.186}
        & \best{15.28}
        & \best{37.00} \\

        \bottomrule
    \end{tabular}
    \vspace{-15pt}
\end{table*}

\noindent\textbf{Dynamic Texture Generation.}
We feed the novel identity texture $\mathcal{T}_\mathrm{id}$ and control poses $\{\mathcal{T}_\mathrm{\bar{N}}^{t}\}_{t\in[1,M]}$ to the $\Phi_{\mathrm{DTG}}$ to generate dynamic textures $\{\mathcal{\hat{T}}^{t}\}_{t\in[1,M]}$.
For arbitrary-length inference, we use sliding-window blending of overlapping VAE latents for temporal consistency.
\par \noindent\textbf{Avatar Rendering.} 
Our $\Phi_\mathrm{GGD}$ decodes $\{\mathcal{T}_\mathrm{\bar{N}}^{t}\}_{t=1}^{M}$ and $\{\hat{\mathcal{T}}^{t}\}_{t=1}^{M}$ into 3D Gaussians, which are rendered with the given camera parameters using the Gaussian rasterizer~\cite{kerbl3Dgaussians}.

\section{Results} \label{sec:results}
\par \noindent\textbf{Datasets.}
We collected a large-scale multi-view human dataset named \datasetname{}, which consists of 58 subjects.
Each subject is captured for $27{,}000$ frames by 100 cameras at 4K resolution.
In terms of motion diversity, our dataset substantially surpasses existing ones.
We isolate two subjects for evaluation.
For the $\Phi_\mathrm{GGD}$, we use 56 subjects from \datasetname{} and 478 subjects from MVHPP~\cite{li2025mvhumannet++} for better identity and clothing generalization.
We use registered texture maps from \datasetname{} to train the $\Phi_\mathrm{DTG}$ because of the temporal smoothness and diversity of identity and motion.
In addition to \datasetname{}, we also evaluate our method on public datasets, DeepCap~\cite{deepcap}, DDC~\cite{habermann2021}, and MetaCap~\cite{sun2024metacap} as reported in Tab.~\ref{tab:dataset_comparison_lhm_mv}. For each dataset, we use two subjects and compute the average metric scores.
\par \noindent\textbf{Implementation Details.}
For texture embedding, we first optimize SMPL-X parameters for each frame using the point cloud and 3D markers, and then refine them by processing 120-frame chunks with 20-frame overlap.
For deformation optimization, the sequence is divided into 20-frame chunks with an overlap of 4 frames.
The texture resolution is set to 256, and we uniformly sample 30 to 32 cameras for texture optimization. 
For the $\Phi_\mathrm{GGD}$, we first train it on MVHPP for 1M steps, and then fine-tune it for another 600k steps using 90\% from \datasetname{} and 10\% from MVHPP at 2K resolution. 
We use AdamW~\cite{loshchilov2017decoupled} with batch size 1 and learning rate $10^{-4}$.
For $\Phi_\mathrm{DTG}$, we use LoRA~\cite{hu2022lora} to fine-tune~\cite{hu2022lora} the 1.3B Wan video diffusion model~\cite{wan2025wan} with batch size 4 and learning rate $10^{-4}$.
\par \noindent\textbf{Evaluation Metrics.}
We use PSNR, SSIM~\cite{wang2004image}, and LPIPS~\cite{zhang2018unreasonable} to evaluate the image reconstruction quality.
To evaluate generative quality at both the image and video levels, we report FID~\cite{heusel2017gans} and FVD~\cite{unterthiner2018towards}.%
\subsection{Comparison} \label{sec:comparison}
\par \noindent\textbf{Competing Methods.} 
\textit{Static methods:}
Initial static avatars are obtained from two input modalities: monocular and multi-view images.
LHM~\cite{qiu2025lhm} is a generalizable method that creates static human avatars from monocular images.
MV denotes reconstructing static Gaussian avatars from multi-view images via gradient descent.
The obtained static avatars are animated in world space using body motion, as described in Sec.~\ref{subsec:preliminaries}.
\textit{Dynamic methods:}
GAS~\cite{lu2025gas} is a video diffusion-based model that first renders a static avatar to 2D images and then generates surface details in image space.
Vid2Avatar-Pro (VAP)~\cite{guo2025vid2avatar} is a regression-based method that maps identity and pose information to 3D Gaussians in texel space.
\par
With methods above, we evaluate how different \textbf{design choices} for \textbf{dynamics embedding} affect dynamic wrinkle generation.
Tab.~\ref{tab:dataset_comparison_lhm_mv} show quantitative comparisons on four datasets, and Fig.~\ref{fig:comparison} visualizes qualitative comparisons on DynaHuman subjects.
Compared to LHM~\cite{qiu2025lhm}, MV leverages more input views for reconstruction, resulting in better performance on both reconstruction and generative metrics.
GAS~\cite{lu2025gas} adds surface dynamics in the image space, breaking both multi-view and identity consistency.
It performs worst in terms of generative metrics.
As a regression-based model, VAP~\cite{guo2025vid2avatar} tends to regress averaged low-frequency signals, such as coarse shape deformation and lighting variations, which leads to strong reconstruction metrics.
However, it struggles to regress high-frequency, motion-dependent surface dynamics and therefore produces overly smooth, nearly uniform wrinkles, resulting in worse generative metrics.
Our method achieves the best generative metrics at both the image and video levels, the second-best reconstruction quality, and consistently better visual results for high-frequency surface dynamics.
Since generation is probabilistic, the generated dynamics may differ from the ground truth, which can lead to high-frequency rendering differences while remaining visually plausible. 
In contrast, regression-based methods such as VAP~\cite{guo2025vid2avatar} tend to conservatively predict low-frequency appearance changes, yielding strong reconstruction metrics but weak explicit surface dynamics.
Therefore, for surface dynamics generation, generative metrics are more suitable than reconstruction metrics, as they better reflect the plausibility of high-frequency motion-dependent details.
\subsection{Ablation Studies} \label{sec:ablation}
\par \noindent\textbf{Human Representation.} 
We evaluate different representations for compressing and reconstructing multi-view human data on one \textbf{training subject}.
We compare two representations: (1) VAP~\cite{guo2025vid2avatar} with static textures (ST), and (2) our optimized dynamic reference textures (RDT) decoded by the \textit{Generalized Gaussian Decoder}.
VAP embeds dynamic information into a regression model, where pose controls guide the prediction of dynamic 3D Gaussians from static identity textures.
While compact, it struggles to reconstruct high-frequency surface dynamics.
In contrast, we embed dynamic information explicitly into optimized dynamic reference textures, which can be accurately decoded to recover the original multi-view data with acceptable storage overhead.
Moreover, these dynamic textures provide a lightweight 2D generative space for training our \textit{Dynamic Texture Generator}, making it easier to synthesize fine-grained motion-dependent dynamics than directly regressing 3D Gaussians.
\begin{table}[t]
\centering
\caption{\textbf{Quantitative Comparison on Human Representation.} 
We compare our human representation with VAP~\cite{guo2025vid2avatar} on Subject59 from the training set. Our \textit{Generalized Gaussian Decoder} (GGD) with optimized reference dynamic textures (RT) achieves better image reconstruction with only a moderate storage increase.
}
\vspace{-5px}
\label{tab:ablation_representation}
\small 
\resizebox{\columnwidth}{!}{
\begin{tabular}{|l|ccc|c|c|}
\hline
\textbf{Representation} & \textbf{PSNR} & \textbf{SSIM} & \textbf{LPIPS}  & \textbf{Tex/Full Storage} \\ 
\hline
ST + Decoder (VAP)    & \second{30.36}     &  \second{0.808}    &  \second{0.160}   & \best{89KB  / 67.6GB}    \\ 
\hline
RDT + Decoder (Ours)   & \best{32.02}      & \best{0.874}     &  \best{0.103} & \second{110MB / 67.6GB}       \\ 
\hline
\end{tabular}
}
\vspace{-5pt}
\end{table}

\begin{table}[t]
\footnotesize 
\centering
\caption{\textbf{Quantitative Ablation on Training Datasets.}
Training with both datasets results in improved generalization and higher reconstruction accuracy.}
\vspace{-5px}
\label{tab:ablation_dataset}
\small 
\scalebox{0.9}{
\begin{tabular}{|l|c|ccc|}
\hline
\textbf{Train Set}  & \textbf{Test Set}  & PSNR  & LPIPS   & FVD   \\ 
\hline
 w/o DynaHuman & DynaHuman     & 28.77        &  0.142          &   33.60  \\ 
 w/o MVHPP & DynaHuman   & \second{29.70}          &  \second{0.124}     &  \second{25.77}     \\ 
\hline
\textbf{Ours}          & DynaHuman  &  \best{29.71}      &  \best{0.122}     &   \best{22.97}  \\
\noalign{\global\arrayrulewidth=1.2pt}\hline
\noalign{\global\arrayrulewidth=0.4pt}
 w/o DynaHuman & MVHPP    & \second{29.45}       &  \best{0.190}           &   -  \\ 
 w/o MVHPP  & MVHPP  & 27.22         &  \second{0.213}        &  -     \\ 
\hline
\textbf{Ours}       & MVHPP    &  \best{29.51}     &  \best{0.190}     &   -  \\
\hline
\end{tabular}
}
\end{table}

\par \noindent\textbf{Influence of Training Dataset.}
The generalization of our $\Phi_\mathrm{GGD}$ depends on the quality and diversity of its training data.
MVHPP~\cite{li2025mvhumannet++} contains diverse subjects but provides fewer motion variations and slightly lower image quality. As a complement, our \datasetname{} offers higher image quality and richer motion diversity but includes fewer subjects.
Joint training on both datasets achieves the best reconstruction performance and generalization (see Tab.~\ref{tab:ablation_dataset}).
\par
\begin{figure}[tb]
    \centering
	\includegraphics[width=0.95\linewidth]{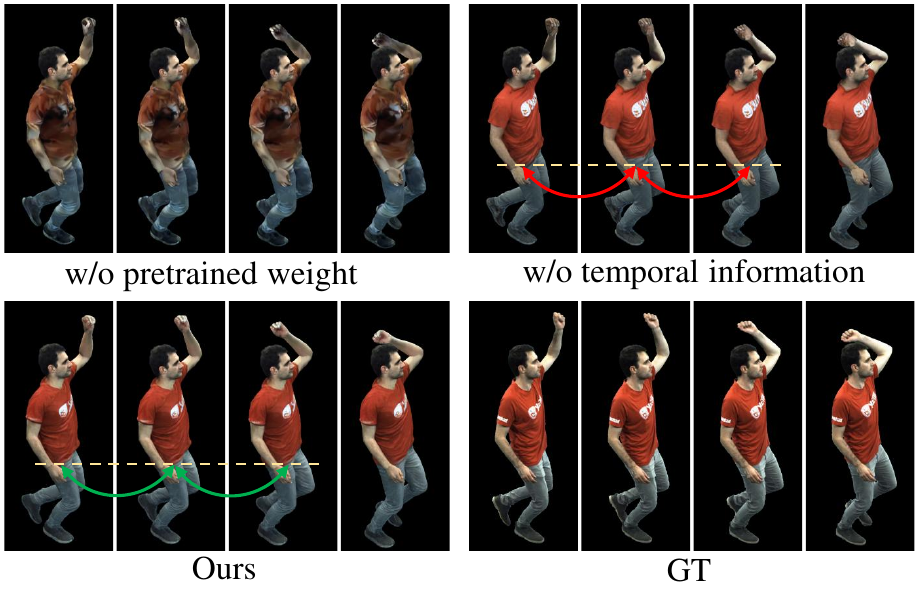}
    \vspace{-5px}
	\caption
	{
	    \textbf{Qualitative Ablation on Dynamics Prior.}
        Training the \textit{Dynamic Texture Generator} from scratch without pretrained diffusion weights yields limited quality.
        Removing the temporal block reduces temporal consistency and causes frame-to-frame variations (red arrows).
        The full model achieves the best generation quality and temporal consistency.
	}
	\label{fig:ablation_prior}
    \vspace{-15pt}
\end{figure}

 \noindent\textbf{Pretrained Video Model Weights.}
Dynamic multi-view training data remains scarce, making pretrained video diffusion priors~\cite{blattmann2023stable} important for learning generalizable surface dynamics.
Training the \textit{Dynamic Texture Generator} without a pre-trained prior (PP) would make convergence and generalization difficult, leading to poor results, as illustrated in Fig.~\ref{fig:ablation_prior} and Tab.~\ref{tab:ablation_prior}.
This underscores the importance of being able to leverage pre-trained priors for dynamics generation and the necessity of our texture representation.
\par \noindent\textbf{Temporal Model vs. Image Model.}
To evaluate temporal modeling, we set the temporal window to 1 for the \textit{Dynamic Texture Generator}, removing temporal information (TI). 
While per-frame quality remains similar, temporal quality degrades significantly, as reflected by the FVD in Tab.~\ref{tab:ablation_prior}. 
Moreover, the model suffers from sudden changes in cloth appearance, as shown by the red arrows and dashed line in Fig.~\ref{fig:ablation_prior}.
\par \noindent\textbf{Generation vs. Reconstruction.}
Given generated textures (Ours) and reference dynamic textures (RDT), we ablate the ability of our full generation pipeline and reconstruction module in Tab.~\ref{tab:ablation_prior} and Fig.~\ref{fig:comparison}.
While our method outperforms baseline dynamic methods, RDT serves as an upper bound by directly using optimized dynamic textures.
\begin{table}
\footnotesize 
\centering
\caption{%
    \textbf{Quantitative Ablation on Dynamics Prior} with Subject71.
    Without pretrained diffusion weights, our \textit{Dynamic Texture Generator} has limited performance.
    Removing the temporal block results in degraded temporal performance.%
}
\vspace{-5px}
\label{tab:ablation_prior}
\scalebox{0.94}{
\begin{tabular}{|l|ccccc|}
\hline
\textbf{Baselines} & PSNR & SSIM & LPIPS & FID   & FVD   \\ 
 \hline
w/o PP                 &  24.70   & 0.700     & 0.240    &  60.42   & 228.21  \\ \hline
w/o TI            & \best{26.61}     &  \best{0.748}   & \second{0.187}   & \second{16.91}   & \second{55.20}   \\ 
\hline
 \textbf{Ours}                      & \best{26.61}     &  \second{0.745}    & \best{0.182}      &  \best{13.84}  &  \best{\textbf{37.74}}   \\ 
 \noalign{\global\arrayrulewidth=1.2pt}\hline
\noalign{\global\arrayrulewidth=0.4pt}
 w/ RDT                      &   29.89  & 0.847    & 0.116     & 8.47   & 20.73    \\ 
\hline
\end{tabular}
}
\vspace{-5pt}
\end{table}

\section{Conclusion} \label{sec:conclusion}
Incorporating dynamics into animatable and generalizable human avatars is crucial for overcoming the uncanny valley and bridging the quality gap between personalized and generalized avatars.
We take an important step in this direction by proposing \methodname{}, a method that transforms off-the-shelf static avatars into dynamic ones.
We demonstrate that our method outperforms both static and dynamic baselines and further highlight the versatility of our approach in multi-modal settings, such as monocular or multi-view image inputs.
Our main technical insight is that dynamics modeling can be formulated as a 2D dynamic texture generation and generalized Gaussian decoding problem, enabling the use of foundational video priors while ensuring multi-view-consistent rendering.
Both properties are essential yet lacking in prior work.
Looking ahead, relaxing the template assumption should be a key research focus to enable modeling of arbitrary clothing and potentially even topological changes.

{\small
\bibliographystyle{ieee_fullname}
\bibliography{mybib}
}

\clearpage
\appendix

In this supplemental material, we present more qualitative (Sec.~\ref{supp:multimodal} and Sec.~\ref{supp:more_results}) and ablative (Sec.~\ref{supp:ablations}) results.
We discuss the drawbacks of big 2D-only method and compare our method with them (Sec.~\ref{supp:wana2}).
Then we provide more details regarding the large-scale DynaHuman dataset (Sec.~\ref{supp:dataset}).
Further, we provide more implementation details regarding texture optimization (Sec.~\ref{supp:optimization}) and network structures (Sec.~\ref{supp:details}).
We perform user study on comparison methods (Sec.~\ref{supp:user_study}).
We discuss the limitations of our method (Sec.~\ref{supp:limitation}).
Lastly, we carefully discuss the potential ethical issues and concerns (Sec.~\ref{supp:ethical}).
\section{Conceptual Comparison of Dynamic Human Representation Paradigms} \label{supp:concept}
\begin{figure*}[tb]
    \centering
	\includegraphics[width=0.92\linewidth]{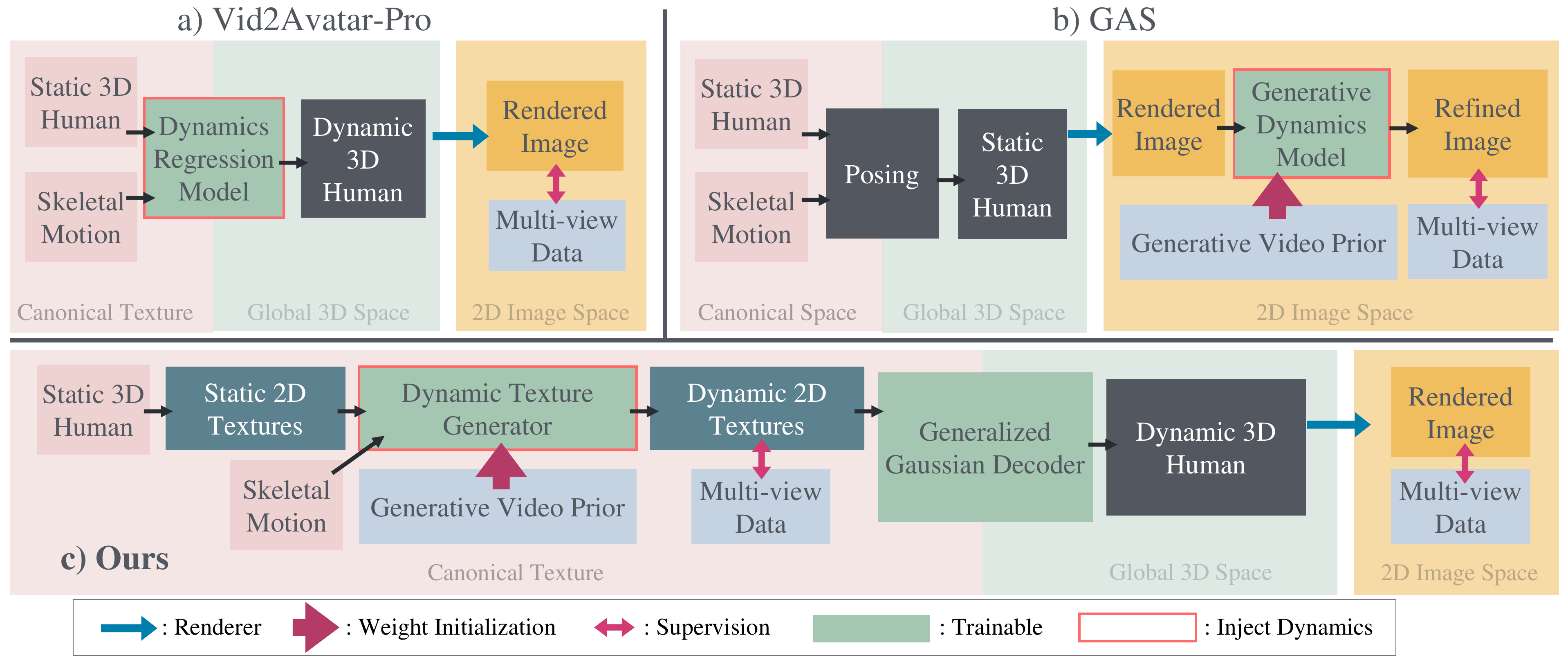}
    \vspace{-5px}
	\caption
	{
	    \textbf{Conceptual Comparison of Generalized and Dynamic Human Representations}.
        a) 
        Regressing 3D human dynamics directly cannot leverage foundational video priors.
        Thus, such methods rely on large-scale (closed-source) multi-view data.
        While regression-based models perform well for static reconstruction, they struggle with dynamic reconstruction.
        b) Leveraging a static rendering as conditioning for a video model allows generative modeling of dynamics.
        However, since the model operates in 2D image space, 3D consistency is not ensured.
        c) Our human representation unites the best of both worlds and enables generative modeling and video prior usage for dynamics while our \textit{Generalized Gaussian Decoder} predicts a dynamic 3D human that is 3D consistent.
	}
    \vspace{-5pt}
	\label{fig:method_comparison}
\end{figure*}

Fig.~\ref{fig:method_comparison} compares three design choices for dynamic human representation. 
Regression-based approaches are effective for static 3D reconstruction, but their extension to dynamic humans remains difficult.
Deterministic regression is not well suited for modeling fine-grained surface dynamics under limited data, as motion-dependent details such as wrinkles are high-frequency and inherently ambiguous: the same identity and pose can correspond to multiple plausible surface states depending on unobserved physical factors. 
As a result, these methods often produce over-smoothed or nearly static dynamics, even when motion conditions are provided. To alleviate this limitation, they typically require large-scale multi-view motion data, which is expensive to collect and often not publicly available.
Another option is to condition a video model on a static human rendering, allowing the model to generate plausible dynamics from learned 2D video priors. However, since the generation is performed in image space, the results do not explicitly guarantee 3D consistency across time and viewpoints. 
Our representation is designed to address this gap. It leverages video priors for dynamic generation while using the Generalized Gaussian Decoder to produce a dynamic 3D human representation that remains consistent in 3D.

\section{Multi-modal Input Results} \label{supp:multimodal}

\begin{figure*}[t]
    \centering
	\includegraphics[width=0.95\linewidth]{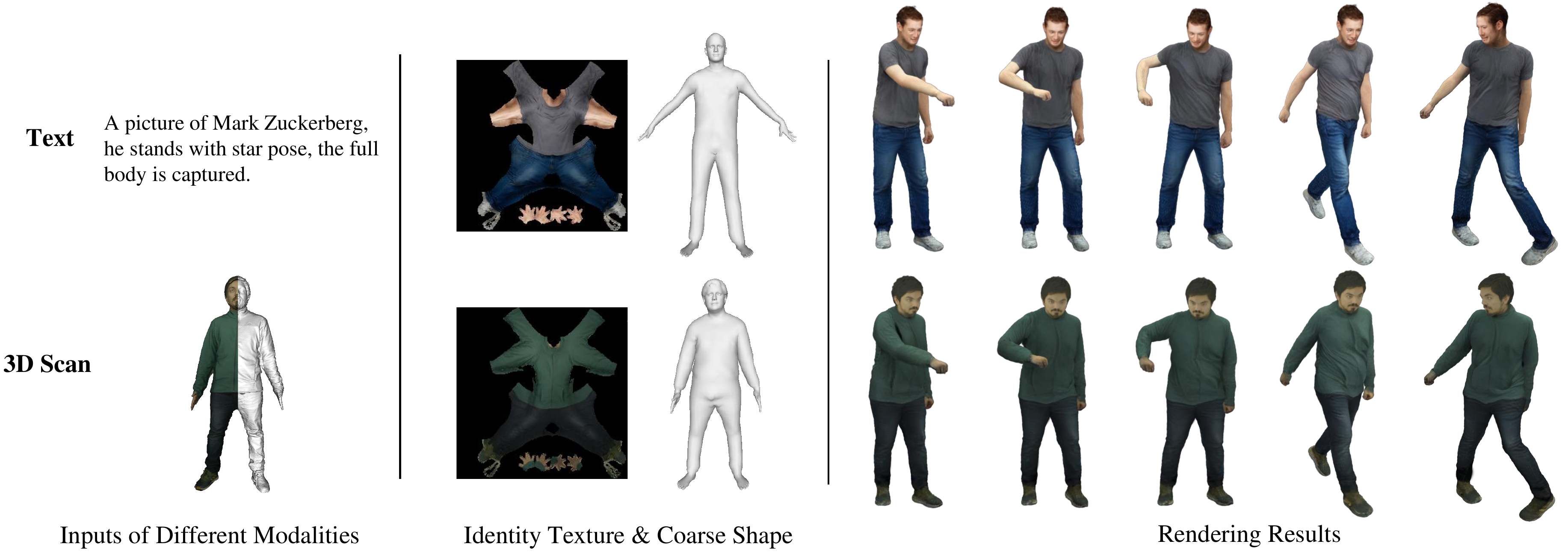}
	\caption
	{
	    \textbf{Multi-modal Avatar Generation}.
        Our method transforms static avatars that are acquired from multi-modal inputs into dynamic ones.
        Here, we show static avatars that are created from text and 3D scans. 
        Next, the static avatar is converted into identity texture and coarse shape. 
        Finally, we can drive the avatar using novel skeletal motions.
        We highlight the different wrinkle patters and dynamics that are induced by the motion and that our method can faithfully generate.
	}
	\label{fig:multimodal}
\end{figure*}

\begin{table*}[t]
  \centering
      \caption{
      \textbf{Dataset Comparison.}
      Statistics are based on the publicly available versions of the datasets (e.g., MVHPP~\cite{li2025mvhumannet++} only releases part of the data). 
      Our dataset matches or surpasses person-specific approaches~\cite{peng2021animatable,habermann2021,sun2025real} in motion diversity and number of views while providing significantly more identities. 
      Compared to large-scale identity datasets~\cite{cheng2023dna,li2025mvhumannet++}, our dataset has significantly higher motion diversity, views, and frames per subject. 
      Besides, we provide high-precision geometry and registered textures.
      Thus, our dataset bridges an important gap in literature to cover surface dynamics.
      }
      \label{tab:dataset}
    \vspace{-5px}
    \small
    \scalebox{0.9}{
    \begin{tabular}{|c|c|c|c|c|c|c|c|c|c|}
    \hline
    \multirow{2}{*}{Dataset}  & Frames & \multirow{2}{*}{Views} & \multirow{2}{*}{Identities}  & Motion  & \multirow{2}{*}{Geometry}  & \multirow{2}{*}{Texture}   & 4K  & Frames\\
      & Per Subject &  &  &  Diversity &  &   & Resolution & (In total)\\
     \hline
     DDC~\cite{habermann2021}  & $\sim$ $26{,}000$ & $\sim$ 90 & 5 & \cmark & \xmark & \xmark & \cmark  &  11.8M \\
     \hline
     DUT~\cite{sun2025real}  & $\sim$ $35{,}000$ & $\sim$ 116 & 2 & \cmark & \cmark & \xmark & \cmark  &  8.1M \\
     \hline
     ZJU~\cite{peng2021neural}  & $\sim$ 500-$2{,}000$	& 24 &	9  & \xmark & \xmark & \xmark & \xmark  &  180K \\
     \hline
     DNA~\cite{cheng2023dna}  & $\sim$ 200	& 60 &	500  & \xmark & \cmark & \xmark & (\cmark)  &  67.5M \\
     \hline
     MVHPP~\cite{li2025mvhumannet++}  & $\sim$ 60	& 16 &	$4{,}500$  & \xmark & \cmark & \xmark & \xmark  &  645.1M \\
     \hline
     \textbf{Ours}  & $27{,}000$ & 100 & 58 & \cmark & \cmark & \cmark & \cmark  &  156.6M \\
    \hline
    \end{tabular}
    }
    \vspace{-10pt}
\end{table*}

We present additional qualitative results in Fig.~\ref{fig:multimodal}, showing that our method can generate dynamic avatars from multi-modal inputs, like text~\cite{dalle} and 3D scans~\cite{teufelgera2025HumanOLAT}. 
Given either textual input or a static 3D scan, we first generate the identity texture and a coarse body shape, represented by an SMPL-X template with personalized offsets.
Through our proposed pipeline, we synthesize clothing dynamics induced by the subject's motion, enriching the static rigged avatar with realistic and expressive motion-aware wrinkle details.
\section{Qualitative Ablations} \label{supp:ablations}
\par
\begin{figure*}[tb]
  \centering
  \includegraphics[width=0.9 \linewidth]{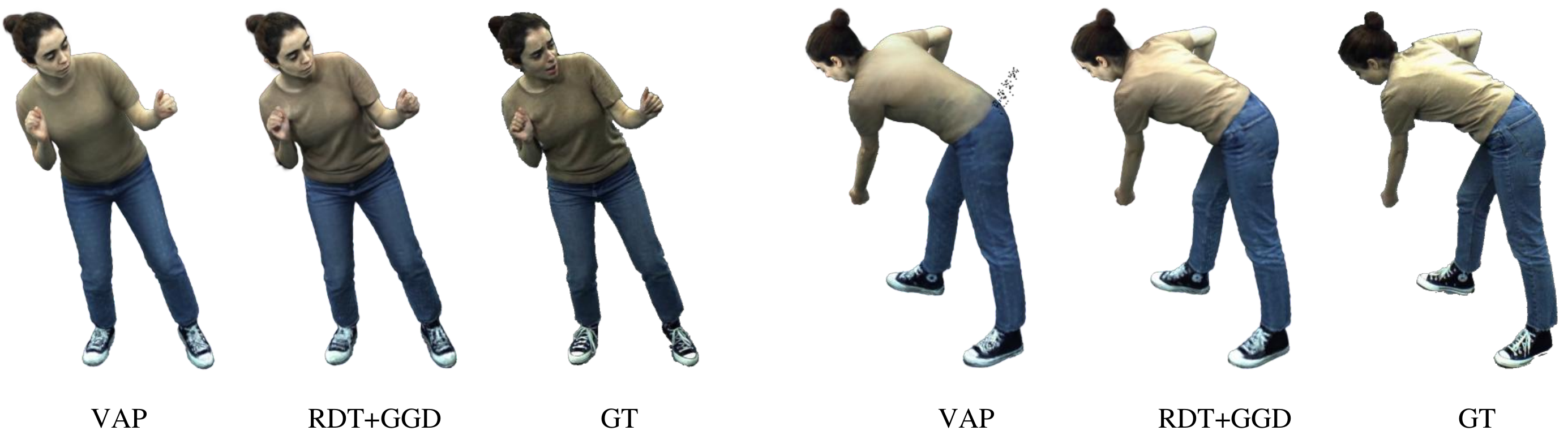}
  \caption{
            \textbf{Representation Ablation.} The qualitative ablation over the avatar representation on the training split. Given the same multi-view videos, compared with the baseline approach, i.e., VAP~\cite{guo2025vid2avatar}, our proposed Generalized Gaussian Decoder (GGD), which takes the reference dynamic texture (RDT) as input, reconstructs the pose-dependent wrinkles more accurately.
        }
    \label{fig:suppl_ablation}
\end{figure*}

 \noindent \textbf{Representation Ablation.}
We compare our dynamic-texture representation against the static-texture representation of VAP~\cite{guo2025vid2avatar}.
Both methods have access to the full multi-view videos of the training subject. However, VAP implicitly encodes dynamics within the regression model through a static identity texture, whereas \methodname{} supplies an explicit per-frame dynamic texture to the \textit{Generalized Gaussian Decoder} (GGD).
As shown in \cref{fig:suppl_ablation}, the GGD faithfully reconstructs fine-grained wrinkles when they are explicitly represented in the dynamic texture. In contrast, the static-texture baseline results in a flat and blurry appearance.

\par
\begin{figure*}[tb]
  \centering
  \includegraphics[width=0.9 \linewidth]{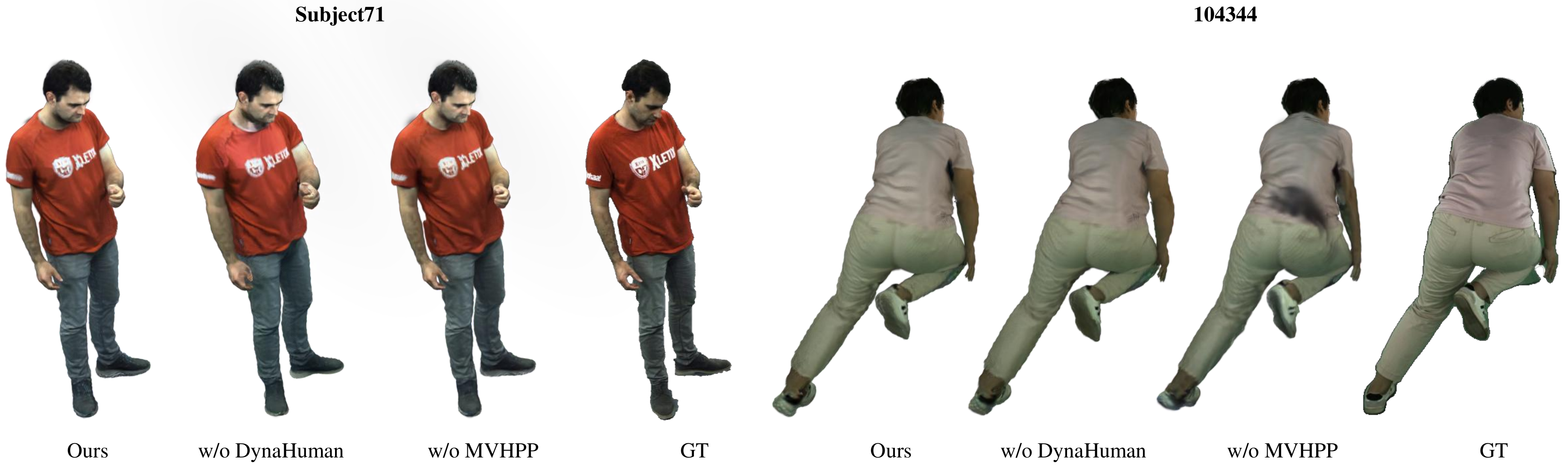}
  \caption{
            \textbf{Dataset Ablation.} The qualitative ablation over the training datasets. 
            Removing the DynaHuman dataset from training leads to uneven lighting, while excluding the MVHPP dataset~\cite{li2025mvhumannet++} from training could cause severe artifacts.
            By combining our DynaHuman dataset and the MVHPP dataset, our Generalized Gaussian Decoder achieves better reconstruction.
        }
    \label{fig:suppl_ablation_dataset}
\end{figure*}

 \noindent \textbf{Training Dataset Ablation.}
Fig.~\ref{fig:suppl_ablation_dataset} presents qualitative ablations analyzing the influence of the training datasets on the \textit{Generalized Gaussian Decoder}.
Removing either DynaHuman or MVHPP~\cite{li2025mvhumannet++}
from training
leads to inaccurate coarse shape estimation, uneven lighting, and ghosting artifacts.
In contrast, leveraging both datasets leads to significantly improved robustness and reconstruction quality.

\section{DynaHuman Dataset} \label{supp:dataset}
\par \noindent \textbf{Capture setup.}
\begin{figure*}[tb]
    \centering
	\includegraphics[width=0.95\linewidth]{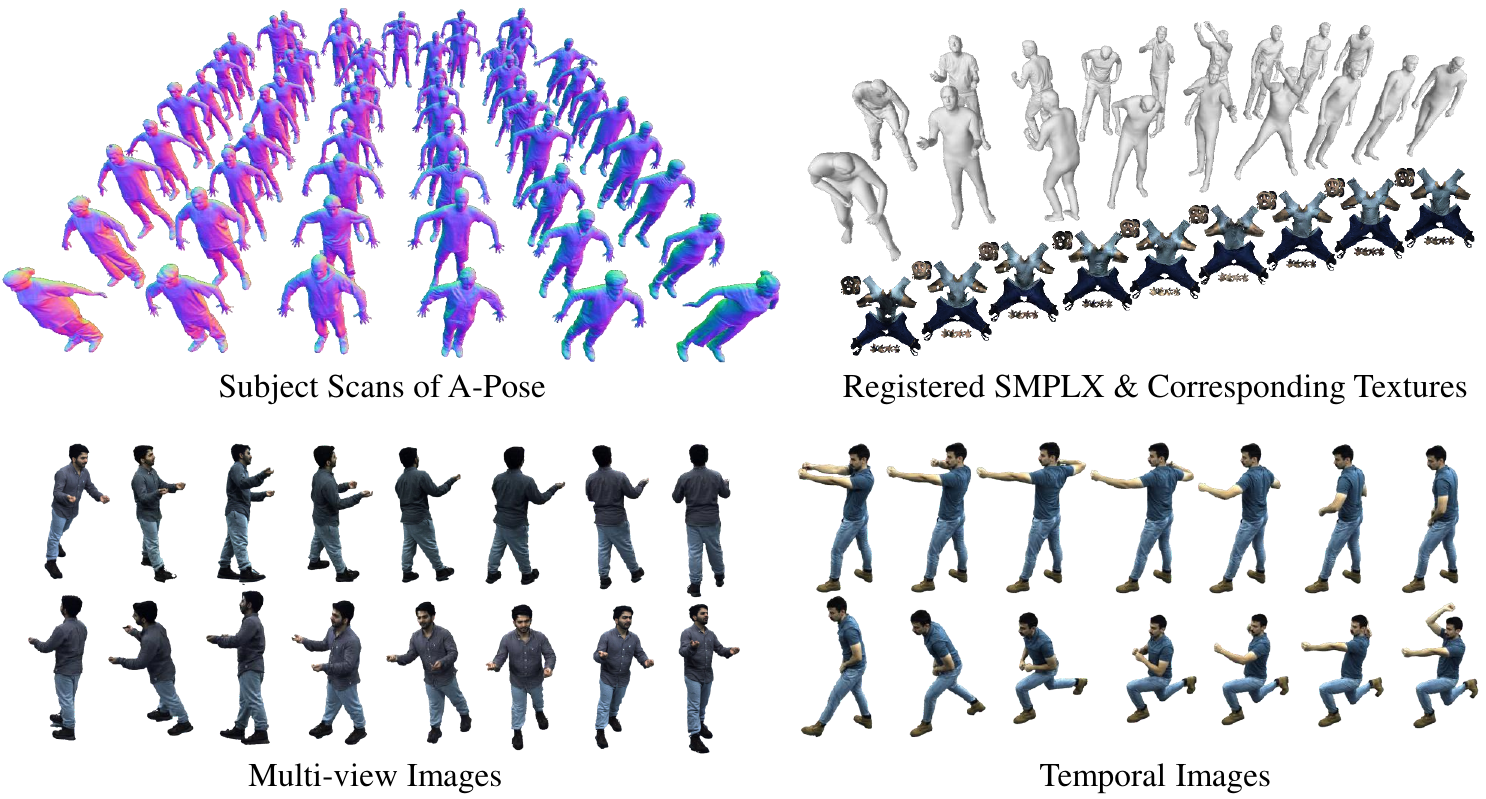}
	\caption
	{
	    \textbf{Dataset Visualization}. For each subject, we provide an A-pose 3D scan, multi-view videos, and registered and deformed SMPL-X as well as dynamic texture annotations.
	}
	\label{fig:dataset}
\end{figure*}

\datasetname{} is captured in a dedicated multi-camera studio equipped with 100 synchronized, calibrated $4K$ cameras arranged in a dome configuration that covers the subject from all elevations.
Notably, the studio is equipped with uniform lighting, which facilitates accurate photometric geometry reconstruction.
As shown in Fig.~\ref{fig:dataset} and Tab.~\ref{tab:dataset},
subjects in \datasetname{} are dressed in diverse clothing styles, such as casual, formal, and sportswear, with varying degrees of tightness, which enhances the diversity of clothing deformation and dynamics.
\par \noindent \textbf{Statistics.}
The dataset comprises 58 subjects, each recorded for approximately 18 minutes ($27{,}000$ frames at $25$ fps) of continuous, natural full-body motion.
Each subject performs motions including locomotion (walking, running, jogging), dynamic whole-body actions (jumping, squatting, turning), and expressive upper-body gestures (arm raises, waving, reaching), deliberately chosen to elicit large, rapid cloth deformations.
Each frame in the recorded multi-view sequence is annotated with SMPL-X tracking, along with ground-truth geometry reconstructed using NeuS2~\cite{neus2}.
Additionally, we provide the corresponding texture maps as described in the main paper.
\par \noindent \textbf{Comparison with existing datasets.}
As summarized in Tab.~1 of the main paper, existing datasets fall into two extremes: DDC~\cite{habermann2021} and DUT~\cite{sun2025real} provide rich clothing dynamics but cover very few identities.
In contrast, MVHumanNet++~\cite{li2025mvhumannet++} and DNA-Rendering~\cite{cheng2023dna} include many subjects, but these subjects mostly perform slow or static motions.
\datasetname{} bridges this gap, featuring both scale (58 subjects) and dynamic diversity, enabling generalizable learning of realistic cloth dynamics.
\section{Texture Embedding via Multi-stage Optimization} \label{supp:optimization}
For each frame, we first obtain point cloud geometry $\mathbf{p}$ and 3D markers $\mathbf{m}$ from multi-view images as illustrated in Sec. 3.3 of the main paper.
Next, we introduce how to register SMPL-X~\cite{SMPL-X:2019} with coarse shape, body motion, and deformations.
\par \noindent \textbf{SMPL-X Tracking.}
We split SMPL-X tracking into three steps.
In the first step, we follow EasyMocap~\cite{easymocap}, which estimates initial SMPL-X parameters from $\mathbf{m}$ as
\begin{align}
&\hat{\mathcal{M}}^{\mathrm{step1}}, \hat{\boldsymbol{\beta}}^{\mathrm{step1}}
=
\arg\min_{\mathcal{M},\,\boldsymbol{\beta}}
\nonumber \\
&\quad
\lambda_m
\left\|
\mathbf{X}\!\left(
\mathbf{V}(\mathcal{M}, \boldsymbol{\beta})
\right)
-
\mathbf{m}
\right\|_2^2
\nonumber \\
&\quad
+
\lambda_{\mathcal{M}}
\sum_{t=2}^{T}
\left\|
\mathcal{M}^t - \mathcal{M}^{t-1}
\right\|_2^2
+
\lambda_\beta \|\boldsymbol{\beta}\|_2^2,
\label{eq:marker_fitting_smooth}
\end{align}
where $\mathbf{X}(\cdot)$ is the marker regressor.
\par
In the second step, we refine the SMPL-X parameters per frame with $\mathbf{p}$ and $\mathbf{m}$ for better alignment, initialized by $\hat{\mathcal{M}}^{\mathrm{step1}}$ and $\hat{\boldsymbol{\beta}}^{\mathrm{step1}}$:
\begin{align}
\hat{\mathcal{M}}^{\mathrm{step2}}, \hat{\boldsymbol{\beta}}^{\mathrm{step2}}
=
\arg\min_{\mathcal{M}^t, \boldsymbol{\beta}^{t} }
\; &
\lambda_m
\left\|
\mathbf{X}\!\left(
\mathbf{V}(\mathcal{M}^t, \boldsymbol{\beta})
\right)
-
\mathbf{m}^t
\right\|_2^2
\nonumber \\
&+
\lambda_p
\left\|
\mathbf{V}(\mathcal{M}^t, \boldsymbol{\beta})
-
\mathbf{p}^t
\right\|_2^2 .
\label{eq:perframe_refine}
\end{align}
\par
In the third step, we smooth the SMPL-X parameters with $\mathbf{p}$ and $\mathbf{m}$, initialized by $\hat{\mathcal{M}}^{\mathrm{step2}}$ and $\hat{\boldsymbol{\beta}}^{\mathrm{step2}}$:
\begin{align}
\hat{\mathcal{M}}^{\mathrm{step3}}
=
\arg\min_{\mathcal{M}}
\; &
\lambda_m
\sum_{t=1}^{T}
\left\|
\mathbf{X}\!\left(
\mathbf{V}(\mathcal{M}^t, \boldsymbol{\beta})
\right)
-
\mathbf{m}^t
\right\|_2^2
\nonumber \\
&+
\lambda_p
\sum_{t=1}^{T}
\left\|
\mathbf{V}(\mathcal{M}^t, \boldsymbol{\beta})
-
\mathbf{p}^t
\right\|_2^2
\nonumber \\
&+
\lambda_{\mathcal{M}}
\sum_{t=2}^{T}
\left\|
\mathcal{M}^t - \mathcal{M}^{t-1}
\right\|_2^2.
\label{eq:temporal_smooth}
\end{align}
\par \noindent \textbf{Deformation Optimization.}
With optimized SMPL-X body motion $\mathcal{M}$ and coarse shape $\boldsymbol{\beta}$, we optimize time-varying cloth displacements $\tilde{\mathbf{d}}_\mathrm{cloth}$:
\begin{align}
&\hat{\tilde{\mathbf{d}}}_\mathrm{cloth} =
\min_{\tilde{\mathbf{d}}_\mathrm{cloth}}
\sum_{t=1}^{T}
\left\|
\mathbf{V}(\mathcal{M}^{t}, \boldsymbol{\beta}, \mathbf{d}_\mathrm{hair}, \tilde{\mathbf{d}}_\mathrm{cloth}^{t}) - \mathbf{p}^{t}
\right\|_2^2
\nonumber \\
&\quad + \sum_{t=1}^{T-1}
\left\|
\mathbf{V}(\mathcal{M}^{t}, \boldsymbol{\beta}, \mathbf{d}_\mathrm{hair}, \tilde{\mathbf{d}}_\mathrm{cloth}^{t})
\right. \nonumber \\
&\qquad\qquad \left.
- \mathbf{V}(\mathcal{M}^{t+1}, \boldsymbol{\beta}, \mathbf{d}_\mathrm{hair}, \tilde{\mathbf{d}}_\mathrm{cloth}^{t+1})
\right\|_2^2,
\end{align}
where $\mathbf{d}_\mathrm{hair}$ comes from the coarse shape of the initial static pose.
\par \noindent \textbf{Texture Optimization.}
Given the deformed human geometry from the previous step, for each frame, we then optimize the texture of the avatar, i.e., unproject the images onto the texture space, using differentiable rasterization.
The exact formulation is provided in Eq.~3 in the main paper.
\section{More Implementation Details} \label{supp:details}
As head modeling is beyond the main scope of body avatar dynamization in this work, we employ a head replacement strategy in VAP~\cite{guo2025vid2avatar} and ours to ensure stable head rendering.
Specifically, we replace the head Gaussian splats of the dynamic avatars with those from the corresponding static avatars.
\par \noindent \textbf{Evaluation.}
For image reconstruction evaluation, we uniformly sample 
every 10 frames from the sequences of the testing subjects and select four camera views that are uniformly placed around the subject.
The metrics are computed at a resolution of $2K$.
For generative quality evaluation, we use $26{,}800$ frames from each test subject sequence and select a camera view facing the subject at the first frame. 
The images are cropped and resized to $512 \times 512$, with the subject centered in each frame.
\par \noindent \textbf{LHM.}
Since LHM~\cite{qiu2025lhm} takes a monocular image as input, we provide a front-facing `A'-pose image and multi-view estimated SMPL-X~\cite{SMPL-X:2019} shape parameters $\boldsymbol{\beta}$ for fair comparison.
We use the official `LHM-500M' checkpoint for all experiments.
We additionally optimize a static Gaussian avatar from LHM outputs with the same number and ordering of points as the 3D methods for head replacement.
\par \noindent \textbf{MV.}
MV denotes reconstructing an initial static human avatar from multi-view images.
Instead of optimizing from scratch, we feed the identity map into our \textit{Generalized Gaussian Decoder} and optimize it for $5k$ steps with a cropped head region loss weight of 1.0 and a regularization weight of 0.01.
\par \noindent \textbf{GAS.}
Since GAS~\cite{lu2025gas} does not provide training code, we use the official checkpoint and perform inference in the `novel pose' mode.
Note that although we do not fine-tune GAS on our \datasetname{} dataset, it has already been trained on several other datasets in addition to MVHPP~\cite{li2025mvhumannet++}, including THuman2.1~\cite{yu2021function4d}, 2K2K~\cite{han2023high}, TikTok~\cite{jafarian2021learning}, and additional internet videos, which are not used for training our approach.
\par \noindent \textbf{VAP.}
Since the source code and training data of VAP~\cite{guo2025vid2avatar} are not publicly available, we re-implement it using the same datasets, network architecture (UNet~\cite{ronneberger2015u}), and training strategy as our \textit{Generalized Gaussian Decoder}.
The main difference between the re-implemented VAP and our \textit{Generalized Gaussian Decoder} lies in the input representation, i.e., a static identity texture versus a dynamic texture.
Our re-implementation slightly differs from their original one -- representing the canonical space in texture space or orthogonal projection space, and predicting Gaussian deltas versus directly predicting Gaussian parameters -- to better evaluate the core difference in representation, i.e., dynamic vs. static textures. 
We highlight that our approach achieves better quality and generalization than VAP while requiring significantly less data, i.e., VAP was originally trained on 1000 dynamic human sequences, which are \textit{not} open source.
\par \noindent \textbf{Ours.}
For the \textit{Generalized Gaussian Decoder}, we adopt a network architecture similar to DUT~\cite{sun2025real}.
The temporal window size of our \textit{Dynamic Texture Generator} is 49, with an overlap of 5 frames during inference.
We fine-tune the video diffusion model for 100k steps on 4 NVIDIA H100 GPUs. We set the LoRA weight to 1.0 for LHM-created avatars and 0.8 for MV-created avatars.
\par \noindent \textbf{RDT+GGD.}
In the reference dynamic texture (RDT) and \textit{Generalized Gaussian Decoder} (GGD) experiment, we use reference textures optimized from multi-view images of the test subjects and feed them into the \textit{Generalized Gaussian Decoder}.
Thus, the model operates in a reconstruction setting, i.e., reconstructing Gaussians from the reference dynamic textures.
\section{More Qualitative Results.} \label{supp:more_results}
\begin{figure*}[tb] %
    \centering
	\includegraphics[width=0.95\linewidth]{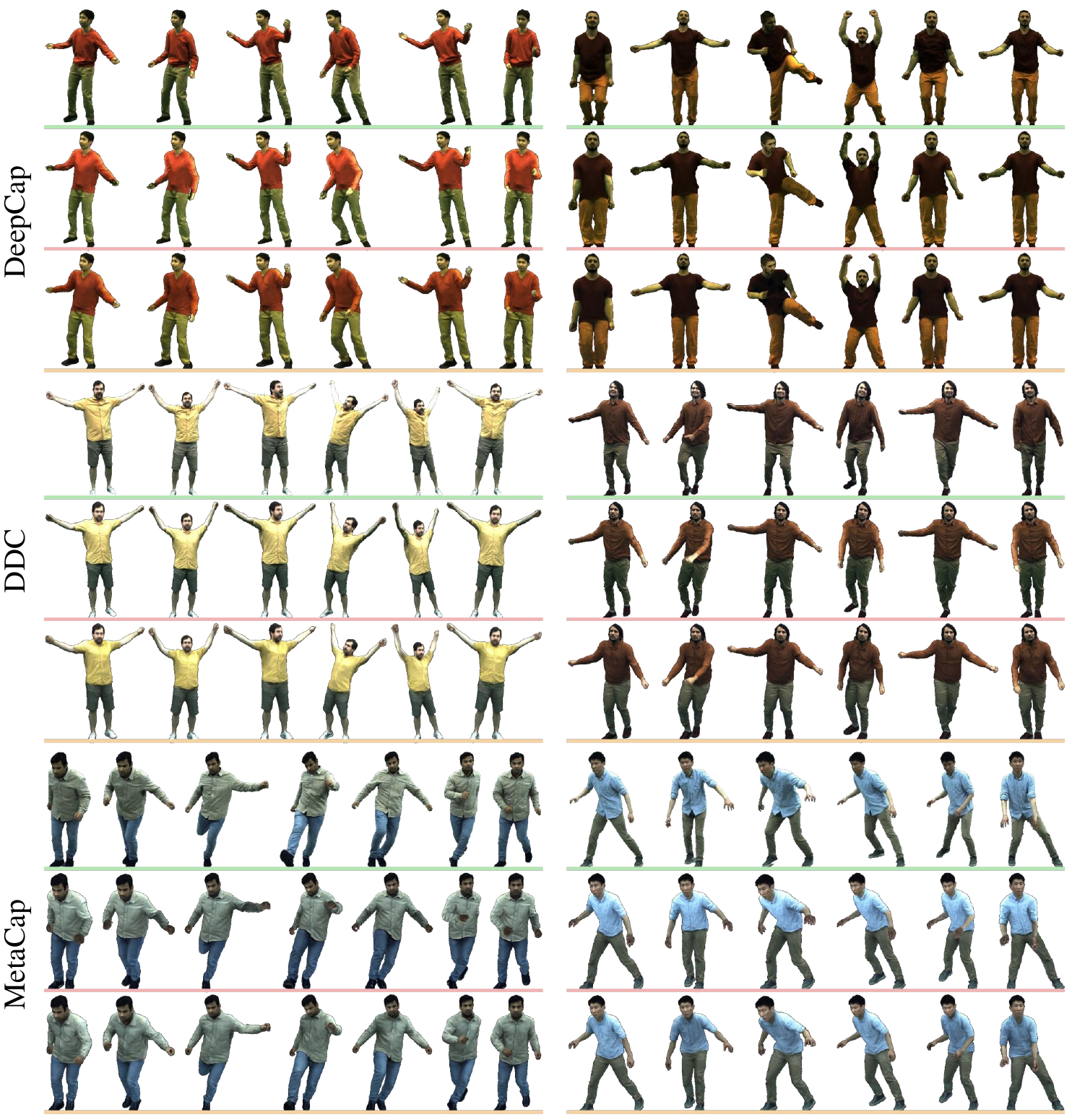}
    \vspace{-5pt}
	\caption
	{
	    \textbf{Qualitative Comparisons on DeepCap, DDC and MetaCap}.
        We show additional qualitative comparisons on public datasets, i.e. DeepCap~\cite{deepcap}, DDC~\cite{habermann2021} and MetaCap~\cite{sun2024metacap}.
        The results are shown in the order of \colorbox[RGB]{208,239,208}{ground truth}, \colorbox[RGB]{246,208,208}{MV (static avatar)}, and \colorbox[RGB]{249,227,195}{our method}.
        Our method successfully generates pose-dependent wrinkles across different motions and clothing types.
	}
	\label{fig:comparison_more}
    \vspace{-5pt}
\end{figure*}

\begin{figure*}[tb] %
    \centering
	\includegraphics[width=0.95\linewidth]{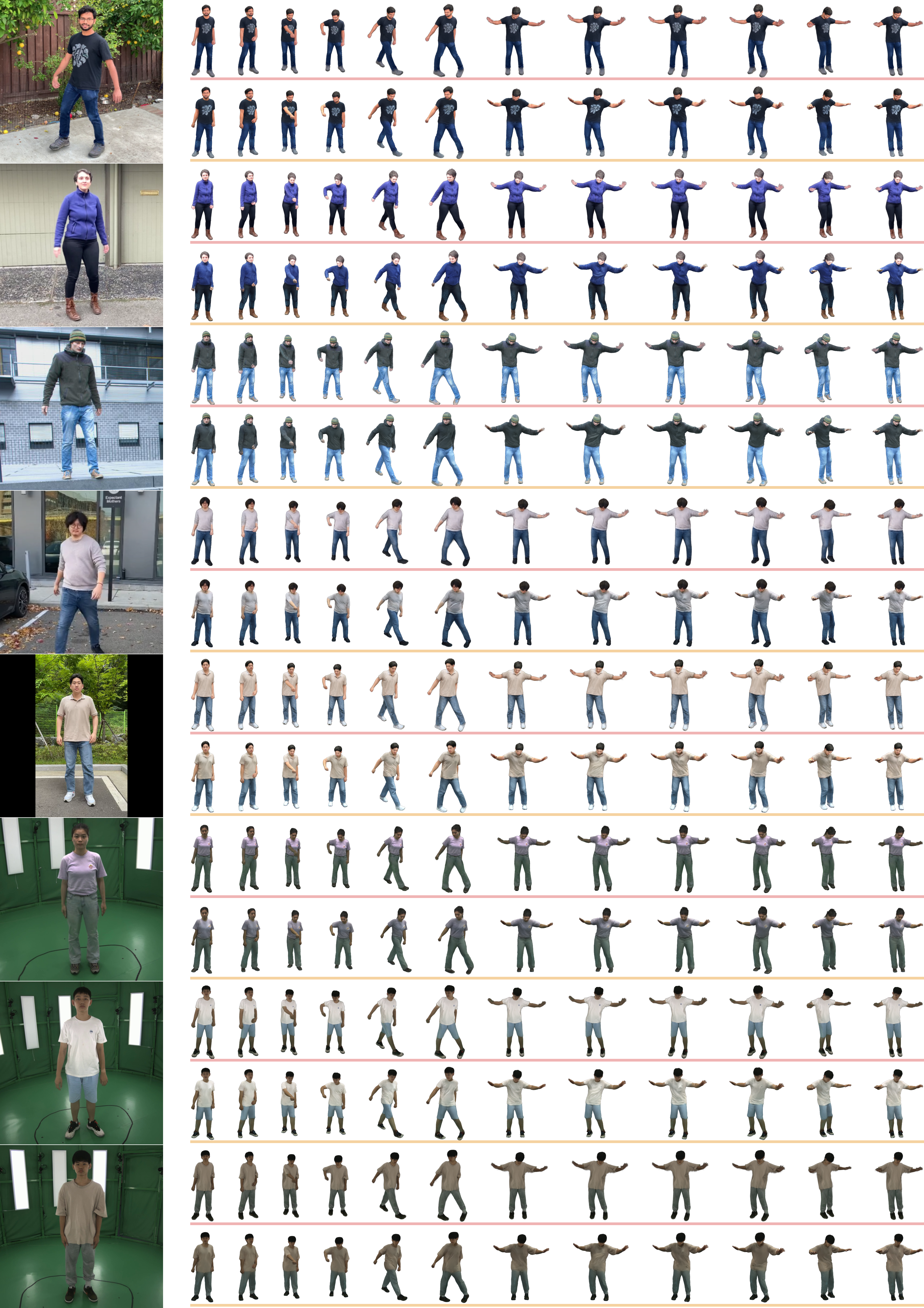}
    \vspace{-5pt}
	\caption
	{
	    \textbf{Qualitative Comparisons on In-the-wild Images}.
        The leftmost image is the input to LHM~\cite{qiu2025lhm}.
        The results are shown in the order of \colorbox[RGB]{246,208,208}{LHM (static avatar)}, and \colorbox[RGB]{249,227,195}{our method}.
         We drive them with the same body motion. Notice that how the wrinkles evolve with the motion and vary across different clothing types.'
	}
	\label{fig:comparison_more_mono}
    \vspace{-5pt}
\end{figure*}

Fig.~\ref{fig:comparison_more} and Fig.~\ref{fig:comparison_more_mono} shows additional qualitative results on DeepCap~\cite{deepcap}, DDC~\cite{habermann2021}, MetaCap~\cite{sun2024metacap} and in-the-wild monocular images.
Our method successfully generates pose-dependent wrinkles across different motions and clothing types.
For in-the-wild subjects, we drive them with the same body motion. The wrinkles evolve with the motion and vary across different clothing types.
\begin{table*}[t]
\centering
\resizebox{0.8\linewidth}{!}{
\begin{tabular}{lcccc}
\hline
 & \begin{tabular}{c}Identity\\Consistency (Q1)\end{tabular}
 & \begin{tabular}{c}Physical\\Plausibility (Q2)\end{tabular}
 & \begin{tabular}{c}View\\Consistency (Q3)\end{tabular}
 & \begin{tabular}{c}Overall\\Score (Q4)\end{tabular} \\
\hline
Static Avatar & \second{3.944/5.0} & \second{3.289/5.0} & \best{4.089/5.0} & \second{3.644/5.0} \\
+GAS          & 1.233/5.0 & 1.289/5.0 & 1.356/5.0 & 1.256/5.0 \\
+VAP          & 3.511/5.0 & 2.622/5.0 & 3.689/5.0 & 2.878/5.0 \\
+Ours         & \best{4.000/5.0} & \best{4.003/5.0} & \second{3.789/5.0} & \best{3.947/5.0} \\
\hline
\end{tabular}
}
\caption{User study questionnaire results. 1 = Poor, 5 = Excellent.}
\label{tab:user_study}
\end{table*}

\section{User Study} \label{supp:user_study}
Beyond qualitative and quantitative evaluations, we conduct a user study to assess human preferences on the perceptual quality of different methods.
A total of 18 participants took part in the study and were asked to compare the results on five subjects according to the following questions:
\begin{itemize}
\item Q1. How well do the methods \textbf{preserve the identity} of the original subject?
\item Q2. How do these methods present \textbf{physically plausible} surface dynamics?
\item Q3. Are these results \textbf{consistent across different views}?
\item Q4. The \textbf{overall quality} considering all technical aspects.
\end{itemize}
As shown in Tab.~\ref{tab:user_study}, our method is preferred in terms of identity preservation, physical plausibility, and overall quality, and achieves the second-best score in view consistency.
Although view consistency is included in the study, it is worth noting that methods based on explicit 3D primitives, including the static avatar baseline, VAP, and ours, are naturally consistent when rendering novel views.

\section{Discussions and Comparisons with Big 2D-only models.} \label{supp:wana2}
\begin{figure*}[tb] %
    \centering
	\includegraphics[width=0.95\linewidth]{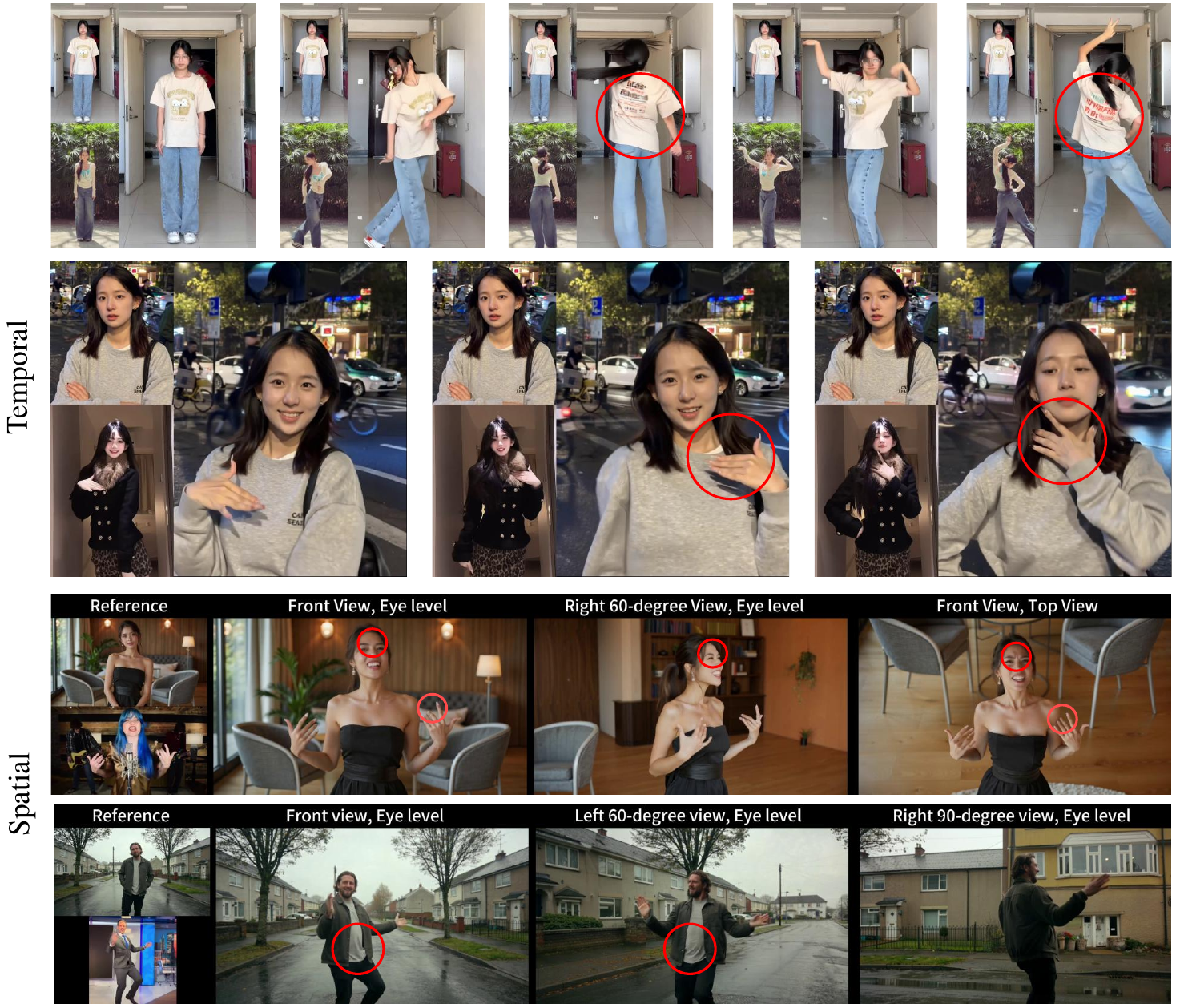}
    \vspace{-5pt}
	\caption
	{
	    \textbf{Failure of Wan-Animate 2}.
        State-of-the-art 2D-only animation models fail to preserve spatial and temporal consistency. 
        Note how the back texture and nail colors change over time (temporal inconsistency), and how facial wrinkles, clothing wrinkles, and nail shapes change across different camera views (spatial inconsistency).
        All results shown above are from the official Wan-Animate 2 results.
	}
	\label{fig:wana2}
    \vspace{-5pt}
\end{figure*}

\begin{figure*}[tb] %
    \centering
	\includegraphics[width=0.95\linewidth]{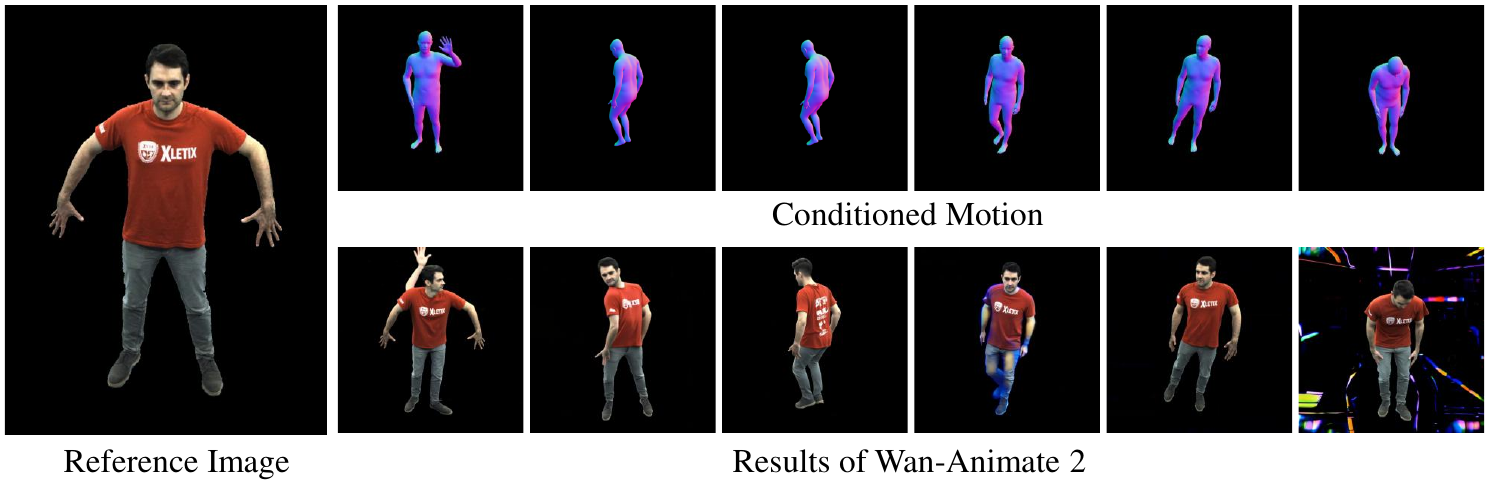}
    \vspace{-5pt}
	\caption
	{
	    \textbf{Additional Failure of Wan-Animate 2}.
        Given a DynaHuman subject, Wan-Animate 2 exhibits several limitations, including incorrect motion embedding, pose-to-appearance ambiguity, inconsistent lighting, distorted body shape, and background artifacts.
	}
	\label{fig:wana2_dyna}
    \vspace{-5pt}
\end{figure*}

Inspired by the success of scaling laws~\cite{kaplan2020scaling}, Wan-Animate 2~\cite{wang2026wananimate2} scales up the 2D-based paradigm exemplified by MagicPose~\cite{chang2023magicpose}, leveraging substantially larger training data and a more advanced diffusion architecture.
Given a reference image and a conditional motion video, these methods generate an animated 2D video of the reference subject following the prescribed motion.
However, they do not employ explicit 3D representations. As a result, their temporal and spatial consistency must be learned implicitly through attention mechanisms~\cite{vaswani2017attention} and large-scale training data, which does not guarantee consistent appearance across time or viewpoints.
Fig.~\ref{fig:wana2} illustrates typical failures in maintaining temporal and spatial consistency.
Fig.~\ref{fig:wana2_dyna} shows additional failure cases of Wan-Animate 2 on a DynaHuman subject, including incorrect motion embedding, pose-to-appearance ambiguity, inconsistent lighting, inaccurate body shape, and background artifacts.
We quantitatively compare our method with Wan-Animate 2~\cite{wang2026wananimate2} in Tab.~\ref{tab:dataset_comparison_wana2}, where our method consistently achieves better performance. Please refer to the main video for qualitative comparisons.

\begin{table}[t]
\centering
\caption{\textbf{Quantitative comparison on DynaHuman.} Here, we compare Wan-Animate 2~\cite{wang2026wananimate2} with our method. Our method achieves consistent better results.
}
\label{tab:dataset_comparison_wana2}
\vspace{-5px}
\scalebox{0.95}{
\begin{tabular}{|l|cc|cc|}
\hline
 & \multicolumn{2}{c|}{\textbf{Subject71}} & \multicolumn{2}{c|}{\textbf{Subject76}} \\
\cline{2-5}
  & FID  & FVD 
  & FID  & FVD \\
\hline
Wan-Animate 2   
& \second{31.83}  & \second{185.09} 
& \second{28.20}  & \second{179.98} \\

\textbf{LHM+Ours} 
& \best{28.20}  & \best{72.66}
& \best{19.65}  & \best{40.78}  \\

\hline
\end{tabular}
}
\vspace{-10pt}
\end{table}

\section{Limitations.} \label{supp:limitation}

Our method effectively transforms a static avatar into a dynamic one, producing realistic renderings with vivid surface dynamics while preserving spatial and temporal consistency.
Nevertheless, several challenges remain:
First, the current pipeline relies on motion tracking and geometry registration, which can be less robust for garments with extreme non-rigid deformation.
Extending generalization to more complex clothing and identifying a suitable canonical space -- potentially a unified 2D representation -- remain future work.
Second, as with most template-based human modeling methods, our representation relies on a fixed mesh template, i.e., SMPL-X, which limits its ability to explicitly model topological changes such as opening a jacket.
While recent work explores Gaussian-based tracking of such changes~\cite{toags}, generalizable tracking across subjects is still challenging.
Third, explicit texture representations may suffer from incomplete observations caused by self-occlusion during texture embedding.

\section{Ethical Discussions} \label{supp:ethical}
Our method enables the generation of vivid, temporally consistent surface dynamics for animatable static human avatars, while supporting a wide range of input modalities. By explicitly modeling fine-grained dynamic textures, it enhances realism, preserves subtle surface details such as wrinkles and cloth deformation, and allows for expressive and controllable avatar animation. These capabilities open up promising applications in areas such as digital communication, virtual reality, telepresence, entertainment, accessibility technologies, and personalized digital humans. In particular, the method can facilitate more natural remote interaction, immersive storytelling, and inclusive human--computer interfaces by bridging the gap between static identity capture and dynamic expressiveness.

As with many powerful generative technologies, the method could be misused. For instance, it may be employed to create fabricated or misleading videos in which an individual's identity is reconstructed from publicly available social media images and subsequently animated according to artificially designed scenarios. Such misuse raises concerns regarding consent and authenticity.

To mitigate these risks, responsible deployment is essential. Appropriate safeguards should therefore be considered prior to practical adoption. These may, for example, include robust identity verification mechanisms, explicit consent requirements for identity reconstruction, or watermarking. By integrating such technical protections along with ethical guidelines and governance measures, the technology can be directed toward socially beneficial use while minimizing potential harm.

\end{document}